\documentclass[sigconf, 9pt]{acmart}

\usepackage{amsmath}
\usepackage{amsfonts}
\usepackage{amsthm}
\usepackage{mathrsfs}
\usepackage{xspace}
\usepackage{xparse}

\usepackage{tabularx}
\usepackage{booktabs}

\usepackage{algorithm}
\usepackage{algpseudocode}
\usepackage{multicol}

\usepackage{accents}

\usepackage[inline]{enumitem}

\usepackage{tikz}
\usetikzlibrary{positioning}

\usepackage{graphicx}
\usepackage{epstopdf}
\usepackage{lipsum}
\usepackage{color, soul}

\usepackage[frozencache,cachedir=.]{minted}

\newtheorem{thm}{Theorem}

\newtheorem{defn}{Definition}
\newtheorem{prob}{Problem}
\newtheorem{prop}{Proposition}
\newtheorem{rem}{Remark}
\newtheorem{assume}{Assumption}

\copyrightyear{2022}
\acmYear{2022}
\setcopyright{acmcopyright}
\acmConference[HSCC '22]{25th ACM International Conference on Hybrid Systems: Computation and Control}{May 4--6, 2022}{Milan, Italy}
\acmBooktitle{25th ACM International Conference on Hybrid Systems: Computation and Control (HSCC '22), May 4--6, 2022, Milan, Italy}
\acmPrice{15.00}
\acmDOI{10.1145/3501710.3519525}
\acmISBN{978-1-4503-9196-2/22/05}

\begin{document}

\title{SOCKS: A Stochastic Optimal Control and Reachability Toolbox Using Kernel Methods}

\author{Adam J. Thorpe}
\orcid{0000-0001-7120-0913}
\affiliation{%
  \department{Electrical \& Computer Eng.}
  \institution{University of New Mexico}
  \streetaddress{P.O. Box 1212}
  \city{Albuquerque}
  \state{New Mexico}
  \country{USA}
}
\email{ajthor@unm.edu}

\author{Meeko M. K. Oishi}
\orcid{0000-0003-3722-8837}
\affiliation{%
  \department{Electrical \& Computer Eng.}
  \institution{University of New Mexico}
  \streetaddress{P.O. Box 1212}
  \city{Albuquerque}
  \state{New Mexico}
  \country{USA}
}
\email{oishi@unm.edu}

\begin{abstract}
We present SOCKS, a data-driven stochastic optimal control toolbox based in kernel methods. SOCKS is a collection of data-driven algorithms that compute approximate solutions to stochastic optimal control problems with arbitrary cost and constraint functions, including stochastic reachability, which seeks to determine the likelihood that a system will reach a desired target set while respecting a set of pre-defined safety constraints. Our approach relies upon a class of machine learning algorithms based in kernel methods, a nonparametric technique which can be used to represent probability distributions in a high-dimensional space of functions known as a reproducing kernel Hilbert space. As a nonparametric technique, kernel methods are inherently data-driven, meaning that they do not place prior assumptions on the system dynamics or the structure of the uncertainty. This makes the toolbox amenable to a wide variety of systems, including those with nonlinear dynamics, black-box elements, and poorly characterized stochastic disturbances. We present the main features of SOCKS and demonstrate its capabilities on several benchmarks.
\end{abstract}

\begin{CCSXML}
<ccs2012>
<concept>
<concept_id>10010147.10010178.10010213.10010214</concept_id>
<concept_desc>Computing methodologies~Computational control theory</concept_desc>
<concept_significance>500</concept_significance>
</concept>
<concept>
<concept_id>10010147.10010257.10010293.10010075</concept_id>
<concept_desc>Computing methodologies~Kernel methods</concept_desc>
<concept_significance>300</concept_significance>
</concept>
<concept>
<concept_id>10003752.10003809.10003716.10011138.10010046</concept_id>
<concept_desc>Theory of computation~Stochastic control and optimization</concept_desc>
<concept_significance>300</concept_significance>
</concept>
</ccs2012>
\end{CCSXML}

\ccsdesc[500]{Computing methodologies~Computational control theory}
\ccsdesc[300]{Computing methodologies~Kernel methods}
\ccsdesc[300]{Theory of computation~Stochastic control and optimization}

\keywords{Stochastic Optimal Control, Machine Learning, Stochastic Reachability}

\maketitle

%%%%%%%%%%%%%%%%%%%%%%%%%%%%%%%%%%%%%%%%%%%%%%%%%%
%%%%%%%%%%%%%%%%%%%%%%%%%%%%%%%%%%%%%%%%%%%%%%%%%%

\section{Introduction}

As modern dynamical systems increasingly incorporate learning enabled components, human-in-the-loop elements, and realistic stochastic disturbances, they become increasingly resistant to traditional controls techniques, and the need for algorithms and tools which can handle such uncertain elements has also grown.
Because of the inherent complexity of these systems, control algorithms based in machine learning are becoming ever more prevalent, and frameworks such as reinforcement learning (RL) and deep neural network controllers have seen widespread popularity in this area--in part because they allow for approximately optimal controller synthesis using a data-driven exploration of the state space and do not rely upon model-based assumptions.
Data-driven control techniques present an attractive approach to stochastic optimal control due to their ability to handle dynamical systems which are resistant to traditional modeling techniques, as well as systems with learning-enabled components and black-box elements.

We present SOCKS, a toolbox for data-driven optimal control based in kernel methods.
The algorithms in SOCKS use a technique known as kernel embeddings of distributions, a nonparametric technique which is rooted in functional analysis and a class of machine learning techniques known collectively as kernel methods \cite{scholkopf2002learning, smola2007hilbert, song2009hilbert}.
Kernel distribution embeddings have been applied to modeling of Markov processes \cite{grunewalder2012modelling, song2010hilbert}, robust optimization \cite{zhu2021kernel}, and statistical inference \cite{song2010nonparametric}.
In addition, these techniques have also been applied to solve stochastic reachability problems \cite{thorpe2020model, thorpe2021approximate}, forward reachability analysis \cite{thorpe2021learning}, and to solving stochastic optimal control problems \cite{thorpe2021stochastic, lever2015modelling}.
Because these techniques are inherently data-driven, SOCKS can accommodate systems with nonlinear dynamics, black-box elements, and arbitrary stochastic disturbances.

Data-driven stochastic optimal control is an active area of research \cite{djeumou2021fly, djeumou2021learning}, and provides a promising avenue for controls problems which suffer from high model complexity or system uncertainty, such as robotic motion planning \cite{kingston2018sampling, marinho2016functional} and model predictive control \cite{rosolia2019sample, rosolia2017learning}.
Recently, approaches using Gaussian processes \cite{deisenroth2009gaussian, rasmussen2006gaussian} and kernel methods \cite{thorpe2021stochastic, lever2015modelling, grunewalder2012modelling} have also been explored.
In SOCKS, we implement the algorithms in \cite{thorpe2021stochastic}, which uses data consisting of observations of the system evolution to compute an implicit approximation of the dynamics in a reproducing kernel Hilbert space (RKHS).
The novelty of the approach in \cite{thorpe2021stochastic} is that it exploits the structure of the RKHS to approximate the stochastic optimal control problem as a linear program that converges in probability to the original problem, and computes an approximately optimal controller without invoking a model-based approach.

The application areas of stochastic optimal control are often strongly motivated by a need for assurances of safety, which presents a need for optimal control techniques which can account for pre-defined safety constraints.
In the reinforcement learning community, this need has led to the development of learning frameworks which enable guided state space exploration strategies \cite[e.g.][]{garcia2015comprehensive, reddy2020learning}, as well as toolsets which implement safety constraint satisfaction as part of the learning loop, such as Safety Gym \cite{ray2019benchmarking}.
SOCKS can also be used to provide assurances of safety using
an established framework known as stochastic reachability \cite{abate2008probabilistic, summers2010verification}, which seeks to determine the likelihood of satisfying a set of pre-specified safety constraints (also called the safety probability).
Numerous toolsets for stochastic reachability have been developed, including \cite{sreachtools, soudjani2014faust, lavaei2020amytiss, shmarov2015probreach, cauchi2019stochy, kwiatkowska2011prism, storm} (see \cite{abate2018arch, abate2019arch, abate2020arch} for a detailed comparison).
In SOCKS, we use the algorithms developed in \cite{thorpe2020model, thorpe2021approximate}, which compute an approximation of the stochastic reachability safety probability using kernel embeddings of distributions.
A recent addition to \mbox{SReachTools} \cite{sreachtools}, presented in \cite{thorpe2020sreachtools}, implements one of the existing stochastic reachability algorithms in SOCKS, but does not consider the stochastic optimal control problem.

Lastly, SOCKS implements an algorithm for forward reachability analysis presented in \cite{thorpe2021learning}.
This technique is useful for analyzing systems with black-box elements, such as deep neural network controllers. Because it employs a data-driven approach, it is agnostic to the structure of the network, and can be used for neural network verification.
Several toolboxes for reachability analysis and verification of deep neural networks have been presented in \cite{nnv, dutta2019sherlock, marabou}.
However, many existing toolsets rely upon prior knowledge of the network structure (such as knowledge of the activation functions), which may not be available without prior knowledge of the system. Because our approach is data-driven, we do not exploit the structure of the system or the network.

The rest of the paper is outlined as follows.
In Section \ref{section: preliminaries}, we describe the class of systems that SOCKS is designed to handle as well as the problems we consider.
In Section \ref{section: distribution embeddings}, we give an overview of the kernel-based techniques used by SOCKS.
Section \ref{section: features} describes the main features of the toolbox.
In Section \ref{section: numerical experiments}, we demonstrate the algorithms in SOCKS on several examples, including a a nonholonomic target-tracking scenario, a realistic satellite rendezvous and docking scenario, a double integrator system to demonstrate stochastic reachability, and on a forward reachable set estimation problem for a neural-network controlled system.
Concluding remarks are presented in Section \ref{section: conclusion}.

%%%%%%%%%%%%%%%%%%%%%%%%%%%%%%%%%%%%%%%%%%%%%%%%%%
%%%%%%%%%%%%%%%%%%%%%%%%%%%%%%%%%%%%%%%%%%%%%%%%%%

\section{Preliminaries}
\label{section: preliminaries}

We use the following notation throughout:
Let $(E, \mathcal{E})$ be an arbitrary measurable space where $\mathcal{E}$ is the $\sigma$-algebra on $E$. If $E$ is topological and $\mathcal{E}$ is the $\sigma$-algebra generated by all open subsets of $E$, then $\mathcal{E}$ is called the Borel $\sigma$-algebra and is denoted $\mathscr{B}_{\mathcal{X}}$.
Let $(\Omega, \mathcal{F}, \mathbb{P})$ be a probability space, where $\mathcal{F}$ is the $\sigma$-algebra on $\Omega$ and $\mathbb{P} : \mathcal{F} \to [0, 1]$ is a probability measure on $(\Omega, \mathcal{F})$.
A measurable function $X : \Omega \to E$ is called an $E$-valued random variable. The image of $\mathbb{P}$ under $X$, $\mathbb{P}(X^{-1} A)$, $A \in \mathcal{E}$, is called the distribution of $X$.
A sequence of $E$-valued random variables $X = \lbrace X_{t} \mid t = 0, 1, \ldots \rbrace$ is called a stochastic process with state space $(E, \mathcal{E})$.

We define a stochastic kernel according to \cite{cinlar2011probability}.

\begin{defn}[Stochastic Kernel]
    Let $(E, \mathcal{E})$ and $(F, \mathcal{F})$ be measurable spaces. A map $\kappa : \mathcal{F} \times E \to [0, 1]$ is a stochastic kernel from $E$ to $F$ if:
    \begin{enumerate*}[mode=unboxed]
        \item
        $x \mapsto \kappa(B \mid x)$ is $\mathcal{E}$-measurable for all $B \in \mathcal{F}$, and
        \item
        $B \mapsto \kappa(B \mid x)$ is a probability measure on $(F, \mathcal{F})$ for every $x \in E$.
    \end{enumerate*}
\end{defn}

We define the indicator function $\boldsymbol{1}_{A} : \mathcal{X} \to \lbrace 0, 1 \rbrace$ for any subset $A \subset E$, such that for any $x \in E$,  $\boldsymbol{1}_{A}(x) = 1$ if $x \in A$ and $\boldsymbol{1}_{A}(x) = 0$ if $x \not\in A$.

%%%%%%%%%%%%%%%%%%%%%%%%%%%%%%%%%%%%%%%%%%%%%%%%%%

\subsection{System Model \& Data}

Consider a discrete-time stochastic system,
\begin{equation}
    \label{eqn: dynamics}
    x_{t+1} = f(x_{t}, u_{t}, w_{t}),
\end{equation}
where $(\mathcal{X}, \mathscr{B}_{\mathcal{X}})$ is a Borel space, $(\mathcal{U}, \mathscr{B}_{\mathcal{U}})$ is a compact Borel space, and $w_{t}$ are independent and identically distributed (i.i.d.) random variables defined on the measurable space $(\mathcal{W}, \mathscr{B}_{\mathcal{W}})$.
The system evolves over a time horizon $t = 0, 1, \ldots, N$, $N \in \mathbb{N}$, from an initial condition $x_{0} \in \mathcal{X}$, which may be drawn from an initial distribution $\mathbb{P}_{0}$ on $\mathcal{X}$, with inputs chosen from a Markov policy $\pi$.

\begin{defn}[Markov Policy]
    A Markov policy is a sequence $\pi = \lbrace \pi_{0}, \pi_{1}, \ldots, \pi_{N-1} \rbrace$, such that for each time $t = 0, 1, \ldots, N-1$, $\pi_{t} : \mathscr{B}_{\mathcal{U}} \times \mathcal{X} \to [0, 1]$ is a stochastic kernel from $\mathcal{X}$ to $\mathcal{U}$.
\end{defn}

We denote the set of all Markov policies as $\Pi$, and for simplicity, we assume the policy is stationary, meaning $\pi_{0} = \pi_{1} = \cdots = \pi_{N-1}$.
We can represent the system in \eqref{eqn: dynamics} as a Markov control process \cite{bertsekas1978stochastic}.

\begin{defn}[Markov Control Process]
    \label{defn: markov control process}
    A Markov control process is a 3-tuple $\mathcal{H} = (\mathcal{X}, \mathcal{U}, Q)$, consisting of
    a Borel space $(\mathcal{X}, \mathscr{B}_{\mathcal{X}})$, a compact Borel space $(\mathcal{U}, \mathscr{B}_{\mathcal{U}})$, and a stochastic kernel $Q : \mathscr{B}_{\mathcal{X}} \times \mathcal{X} \times \mathcal{U} \to [0, 1]$ from $\mathcal{X} \times \mathcal{U}$ to $\mathcal{X}$.
\end{defn}

We consider the case where the stochastic kernel $Q$ is unknown, meaning we have no prior knowledge of the statistical features of $Q(\cdot \mid x, u)$ or the dynamics in \eqref{eqn: dynamics}, but assume that a sample $\mathcal{S}$ collected i.i.d. from $Q$ is available.
We make this scenario more explicit via the following assumptions.

\begin{assume}
    \label{assume: stochastic kernel is unknown}
    The stochastic kernel $Q$ is unknown.
\end{assume}

\begin{assume}
    \label{assume: sample available}
    A sample $\mathcal{S}$ of size $M \in \mathbb{N}$ taken i.i.d. from $Q$ is available,
    \begin{equation}
        \label{eqn: sample}
        \mathcal{S} = \lbrace (x_{1}, u_{1}, y_{1}), \ldots, (x_{M}, u_{M}, y_{M}) \rbrace,
    \end{equation}
    where $x_{i}$ and $u_{i}$ are randomly sampled from a probability distribution on $\mathcal{X} \times \mathcal{U}$ and $y_{i} \sim Q(\cdot \mid x_{i}, u_{i})$.
\end{assume}

%%%%%%%%%%%%%%%%%%%%%%%%%%%%%%%%%%%%%%%%%%%%%%%%%%

\subsection{Problem Definitions}

%%%%%%%%%%%%%%%%%%%%%%%%%%%%%%%%%%%%%%%%%%%%%%%%%%

\subsubsection{Stochastic Optimal Control}

Consider the following stochastic optimal control problem, which seeks to minimize an arbitrary, bounded cost function subject to a set of constraints.

\begin{prob}[Stochastic Optimal Control]
    \label{prob: optimal control problem}
    Let $\mathcal{H}$ be a Markov control process as in Definition \ref{defn: markov control process}, and define the functions $f_{0} : \mathcal{X} \times \mathcal{U} \to \mathbb{R}$, called the \emph{objective} or \emph{cost} function and $f_{i} : \mathcal{X} \times \mathcal{U} \to \mathbb{R}$, $i = 1, \ldots, p$, called the \emph{constraints}.
    We seek a policy $\pi \in \Pi$ that minimizes the following optimization problem:
    \begin{subequations}
        \label{eqn: optimal control problem}
        \begin{align}
            \label{eqn: optimal control problem objective}
            \min_{\pi} \quad & \int_{\mathcal{U}} \int_{\mathcal{X}} f_{0}(y, v) Q(\mathrm{d} y \mid x, v) \pi(\mathrm{d} v \mid x) \\
            \label{eqn: optimal control problem constraints}
            \textnormal{s.t.} \quad & \int_{\mathcal{U}} \int_{\mathcal{X}} f_{i}(y, v) Q(\mathrm{d} y \mid x, v) \pi(\mathrm{d} v \mid x) \leq 0, i = 1, \ldots, p
        \end{align}
    \end{subequations}
\end{prob}

We impose the following mild simplifying assumption which allows us to separate the cost with respect to $x$ and $u$.

\begin{assume}
    \label{assume: cost decomposed}
    The cost and constraint functions $f_{i} : \mathcal{X} \times \mathcal{U} \to \mathbb{R}$, $i = 0, 1, \ldots, p$, can be decomposed as:
    \begin{equation}
        f_{i}(x_{t}, u_{t}) = f_{i}^{x}(x_{t}) + f_{i}^{u}(u_{t}).
    \end{equation}
\end{assume}

Several commonly-known cost functions obey this assumption, such as the quadratic LQR cost function.

The primary difficulty in solving Problem \ref{prob: optimal control problem} is due to Assumption \ref{assume: stochastic kernel is unknown}, and also because we seek a \emph{distribution} $\pi$ which minimizes the objective.
Because $Q$ is unknown, the integral with respect to $Q$ in \eqref{eqn: optimal control problem} is intractable.
Thus, we form an approximation of Problem \ref{prob: optimal control problem} by computing an empirical approximation of the integral operator with respect to $Q$ using the sample $\mathcal{S}$.
Following \cite{thorpe2021stochastic}, we can view this as a learning problem by embedding the integral operator as an element in a high-dimensional space of functions known as a reproducing kernel Hilbert space.
Details regarding the kernel-based stochastic optimal control method are provided in \cite{thorpe2021stochastic} and in Appendix \ref{appendix: stochastic optimal control}.

%%%%%%%%%%%%%%%%%%%%%%%%%%%%%%%%%%%%%%%%%%%%%%%%%%

\subsubsection{Backward Stochastic Reachability}

We also consider a special case of the stochastic optimal control problem in \eqref{eqn: optimal control problem}, known as the terminal-hitting time stochastic reachability problem. As defined in \cite{summers2010verification}, the goal is to compute a policy $\pi \in \Pi$ that maximizes the likelihood that a system $\mathcal{H}$ will remain within a pre-defined safe set $\mathcal{K} \subseteq \mathscr{B}_{\mathcal{X}}$ for all time $t < N$, and reach some target set $\mathcal{T} \subseteq \mathscr{B}_{\mathcal{X}}$ at time $t = N$.
We define the \emph{safety probability} as:
\begin{equation}
    \label{eqn: safety probabilities}
    r_{x_{0}}^{\pi}(\mathcal{K}, \mathcal{T}) = \mathbb{P}_{x_{0}}^{\pi} \lbrace x_{N} \in \mathcal{T} \land x_{i} \in \mathcal{K}, \forall i = 0, 1, \ldots, N-1 \rbrace
\end{equation}
The solution to the stochastic reachability problem is typically formulated as a dynamic program using indicator functions.
Define the value functions $V_{t}^{*} : \mathcal{X} \to [0, 1]$, $k = 0, 1, \ldots, N$ by the backward recursion,
\begin{subequations}
\label{eqn: stochastic reachability backward recursion}
\begin{align}
    V_{N}^{*}(x) &= \boldsymbol{1}_{\mathcal{T}}(x), \\
    \label{eqn: stochastic reachability value function}
    V_{t}^{*}(x) &= \sup_{\pi \in \Pi}
    \boldsymbol{1}_{\mathcal{K}}(x) \int_{\mathcal{X}} V_{t+1}^{*}(y) Q(\mathrm{d} y \mid x, v) \pi(\mathrm{d} v \mid x),
\end{align}
\end{subequations}
where $x \in \mathcal{X}$. Then $V_{0}^{*}(x_{0}) = \sup_{\pi \in \Pi} r_{x_{0}}^{\pi} (\mathcal{K}, \mathcal{T})$.

\begin{prob}[Terminal-Hitting Time Problem]
    \label{prob: stochastic reachability problem}
    We seek to compute an approximation of the policy $\pi \in \Pi$ that maximizes the safety probabilities in \eqref{eqn: safety probabilities}, and converges in probability to the true solution, $\pi^{*} = \arg \sup_{\pi \in \Pi} r_{x_{0}}^{\pi}(\mathcal{K}, \mathcal{T})$.
\end{prob}

Similar to Problem \ref{prob: optimal control problem}, the backward recursion in \eqref{eqn: stochastic reachability backward recursion} is intractable due to Assumption \ref{assume: stochastic kernel is unknown}.
We can use the same technique as Problem \ref{prob: optimal control problem} in order to approximate the value functions in \eqref{eqn: stochastic reachability backward recursion}, and thereby obtain an approximation of the safety probabilities in \eqref{eqn: safety probabilities}.

\begin{rem}
    \label{rem: stochastic reachability problems}
    We note that our toolbox can be used to solve other stochastic reachability problems, including the first-hitting time problem as defined in \cite{summers2010verification} and the max and multiplicative problems defined in \cite{abate2008probabilistic}.
    We focus on the terminal-hitting time problem in the current work for simplicity.
\end{rem}

%%%%%%%%%%%%%%%%%%%%%%%%%%%%%%%%%%%%%%%%%%%%%%%%%%

\subsubsection{Forward Stochastic Reachability}

The forward reachable set $\mathscr{F}$ is defined as the set of all states that the system in \eqref{eqn: dynamics} can reach after $N$ time steps from an initial condition $x_{0} \in \mathcal{X}$. As shown in \cite{thorpe2021learning}, we can view the problem of estimating the forward reachable set $\mathscr{F}$ as a support estimation problem, where the support is the smallest closed set $\mathscr{F} \subset \mathcal{X}$ such that $\mathbb{P}_{N}(x_{N} \in \mathscr{F}) = 1$, where $x_{N}$ is a random variable representing the state of the system at time $N$ and $\mathbb{P}_{N}$ is the distribution of $x_{N}$.

\begin{prob}[Forward Reachability]
    \label{prob: forward reachability problem}
    We seek to determine the support of $\mathbb{P}_{N}$, the state distribution over $(\mathcal{X}, \mathscr{B}_{\mathcal{X}})$ after $N$ time steps.
\end{prob}

We formulate Problem \ref{prob: forward reachability problem} as learning a classifier $F$, where
\begin{equation}
    \mathscr{F} = \lbrace x \in \mathcal{X} \mid F(x) = 1 \rbrace.
\end{equation}
The difficulty in computing the reachable set classifier is due to the fact that the dynamics in \eqref{eqn: dynamics} and the stochastic kernel is unknown (by Assumption \ref{assume: stochastic kernel is unknown}). Thus, we approximate the classifier function $F$ as an element in an RKHS, and use a sample collected from $\mathbb{P}_{N}$ in order to estimate $F$.

%%%%%%%%%%%%%%%%%%%%%%%%%%%%%%%%%%%%%%%%%%%%%%%%%%
%%%%%%%%%%%%%%%%%%%%%%%%%%%%%%%%%%%%%%%%%%%%%%%%%%

\section{Embedding Distributions in a Hilbert Space of Functions}
\label{section: distribution embeddings}

In this section, we provide an overview of the machine learning techniques used by SOCKS to solve Problems \ref{prob: optimal control problem} and \ref{prob: stochastic reachability problem}. Details are provided in \cite{thorpe2021stochastic, thorpe2020model, thorpe2020sreachtools}. The method used to solve Problem \ref{prob: forward reachability problem} is based in a similar framework, with additional details in \cite{thorpe2021learning}.

Let $k : \mathcal{X} \times \mathcal{X} \to \mathbb{R}$ be a positive definite kernel function.

\begin{defn}[Positive Definite Kernel]
    A kernel $k$ is called positive definite if for all $n \in \mathbb{N}$, $x_{1}, \ldots, x_{n} \in \mathcal{X}$, and $\alpha_{1}, \ldots, \alpha_{n} \in \mathbb{R}$, $\sum_{i=1}^{n} \sum_{j=1}^{n} \alpha_{i} \alpha_{j} k(x_{i}, x_{j}) \geq 0$.
\end{defn}

Let $\mathscr{H}$ denote a Hilbert space of functions from $\mathcal{X}$ to $\mathbb{R}$, equipped with the inner product $\langle \cdot, \cdot \rangle_{\mathscr{H}}$.

\begin{defn}[RKHS, {\cite{aronszajn1950theory}}]
    \label{defn: rkhs}
    A Hilbert space $\mathscr{H}$ of functions $\mathcal{X} \to \mathbb{R}$ is called a reproducing kernel Hilbert space if there exists a positive definite kernel $k$ called the \emph{reproducing kernel}, such that the following properties hold:
    \begin{enumerate}%[mode=unboxed]
        \item
        $k(x, \cdot) \in \mathscr{H}$ for all $x \in \mathcal{X}$, and
        \item
        $f(x) = \langle f, k(x, \cdot) \rangle_{\mathscr{H}}$ for all $f \in \mathscr{H}$ and $x \in \mathcal{X}$.
    \end{enumerate}
\end{defn}

Alternatively, by the Moore-Aronszajn theorem \cite{aronszajn1950theory}, for any positive definite kernel function $k$, there exists a unique RKHS with reproducing kernel $k$. For instance, a commonly-used kernel function is the Gaussian RBF kernel $k(x, x^{\prime}) = \exp(-\lVert x - x^{\prime} \rVert_{2}^{2}/(2 \sigma^{2}))$, $\sigma > 0$. We define the reproducing kernel $k : \mathcal{X} \times \mathcal{X} \to \mathbb{R}$ on $\mathcal{X}$ with the associated RKHS $\mathscr{H}$ and the kernel $l : \mathcal{U} \times \mathcal{U} \to \mathbb{R}$ on $\mathcal{U}$.

The second property in Definition \ref{defn: rkhs} is called the \emph{reproducing property} and is key to our approach.
In short, it allows us to evaluate any function in $\mathscr{H}$ as a Hilbert space inner product.
We use this property to evaluate the integral terms in Problems \ref{prob: optimal control problem} and \ref{prob: stochastic reachability problem} by embedding the integral operator with respect to the stochastic kernel $Q$ as an element in an RKHS.

We now define the following:

\begin{defn}
    A kernel $k$ is called bounded if for some positive constant $\rho > 0$,
    \begin{equation}
        \sup_{x \in \mathcal{X}} \sqrt{k(x, x)} \leq \rho < \infty.
    \end{equation}
\end{defn}

In order to embed a distribution in $\mathscr{H}$, we make the following mild assumption:

\begin{assume}
    \label{assume: kernel bounded and measurable}
    The kernel $k$ is bounded and measurable with respect to $\mathcal{X}$.
\end{assume}

For every $(x, u) \in \mathcal{X} \times \mathcal{U}$, $Q(\cdot \mid x, u)$ is a probability measure on $\mathcal{X}$.
We denote by $\mathscr{P}$ the set of probability measures on $\mathcal{X}$ conditioned on $\mathcal{X} \times \mathcal{U}$, of which the probability measures $Q(\cdot \mid x, u)$ generated by $Q$ are a part.
If the following necessary and sufficient condition is satisfied,
\begin{equation}
    \int_{\mathcal{X}} \sqrt{k(y, y)} Q(\mathrm{d} y \mid x, u) < \infty,
\end{equation}
which holds due to Assumption \ref{assume: kernel bounded and measurable}, then there exists an element $m(x, u) \in \mathscr{H}$ called the \emph{kernel distribution embedding},
which is a mapping,
\begin{align}
    \label{eqn: conditional distribution embedding}
    \begin{split}
        m : \mathscr{P} & \to \mathscr{H}, \\
        Q(\cdot \mid x, u) & \mapsto \int_{\mathcal{X}} k(y, \cdot) Q(\mathrm{d} y \mid x, u).
    \end{split}
\end{align}
Then by the reproducing property, we can evaluate the integral of any function $f \in \mathscr{H}$ as an RKHS inner product,
\begin{align}
    \langle f, m(x, u) \rangle_{\mathscr{H}} &= \biggl\langle f, \int_{\mathcal{X}} k(y, \cdot) Q(\mathrm{d} y \mid x, u) \biggr\rangle_{\mathscr{H}} \\
    &= \int_{\mathcal{X}} \langle f, k(y, \cdot) \rangle_{\mathscr{H}} Q(\mathrm{d} y \mid x, u) \\
    &= \int_{\mathcal{X}} f(y) Q(\mathrm{d} y \mid x, u).
\end{align}

However, in practice, we do not have access to the true embedding $m(x, u)$ due to Assumption \ref{assume: stochastic kernel is unknown}. Thus, we seek an empirical estimate $\hat{m}(x, u)$ of $m(x, u)$ computed using the sample $\mathcal{S}$.

%%%%%%%%%%%%%%%%%%%%%%%%%%%%%%%%%%%%%%%%%%%%%%%%%%

\subsection{Empirical Distribution Embeddings}

Following \cite{grunewalder2012conditional}, we can compute an estimate $\hat{m}(x, u)$ using $\mathcal{S}$ as the solution to the following regularized least-squares problem:
\begin{equation}
    \label{eqn: regularized least squares}
    \hat{m} = \arg \min_{f \in \mathscr{Q}} \frac{1}{M} \sum_{i=1}^{M} \lVert k(y_{i}, \cdot) - f(x_{i}, u_{i}) \rVert_{\mathscr{H}}^{2} + \lambda \lVert f \rVert_{\mathscr{Q}}^{2},
\end{equation}
where $\mathscr{Q}$ is a vector-valued RKHS \cite{micchelli2005learning}, and $\lambda > 0$ is the regularization parameter.
According to \cite{micchelli2005learning}, by the representer theorem, the solution to \eqref{eqn: regularized least squares} has the following form:
\begin{equation}
    \label{eqn: regularized least squares form}
    \hat{m} = \sum_{i=1}^{M} \alpha_{i} k(x_{i}, \cdot) l(u_{i}, \cdot),
\end{equation}
where $\alpha \in \mathbb{R}^{M}$ is a vector of real-valued coefficients. By substituting \eqref{eqn: regularized least squares form} into \eqref{eqn: regularized least squares} and taking the derivative with respect to $\alpha$, we obtain the following closed-form solution:
\begin{equation}
    \label{eqn: conditional distribution embedding estimate}
    \hat{m}(x, u) = \Phi^{\top} (\Psi \Psi^{\top} + \lambda M I)^{-1} \Psi k(x, \cdot) l(u, \cdot),
\end{equation}
where $\Phi$ and $\Psi$ are called \emph{feature vectors} with elements $\Phi_{i} = k(y_{i}, \cdot)$ and $\Psi_{i} = k(x_{i}, \cdot) l(u_{i}, \cdot)$.

For simplicity, let $W \coloneqq (\Psi \Psi^{\top} + \lambda M I)^{-1}$ and define
\begin{equation}
    \label{eqn: beta}
    \beta(x, u) \coloneqq W \Psi k(x, \cdot) l(u, \cdot),
\end{equation}
such that $\hat{m}(x, u) = \Phi^{\top} \beta(x, u)$.
Then by the reproducing property, for any function $f \in \mathscr{H}$, we can approximate the integral of $f$ with respect to $Q(\cdot \mid x, u)$ as an RKHS inner product:
\begin{equation}
    \langle f, \hat{m}(x, u) \rangle_{\mathscr{H}} = \boldsymbol{f}^{\top} \beta(x, u) \approx \int_{\mathcal{X}} f(y) Q(\mathrm{d} y \mid x, u),
\end{equation}
where $\boldsymbol{f}$ is a vector with elements $\boldsymbol{f}_{i} = f(y_{i})$.

In addition, the empirical estimate $\hat{m}(x, u)$ converges in probability to the true embedding $m(x, u)$ as the sample size $M$ increases and the regularization parameter $\lambda$ is decreased at an appropriate rate \cite[see][]{grunewalder2012conditional, song2009hilbert}. Additional details regarding the convergence properties of the embedding are provided in Appendix \ref{section: stability and convergence}.

%%%%%%%%%%%%%%%%%%%%%%%%%%%%%%%%%%%%%%%%%%%%%%%%%%
%%%%%%%%%%%%%%%%%%%%%%%%%%%%%%%%%%%%%%%%%%%%%%%%%%

\section{Features}
\label{section: features}

We implemented SOCKS in Python, which has several available libraries for machine learning and reinforcement learning, such as Tensorflow \cite{tensorflow2015-whitepaper}, Keras \cite{chollet2015keras}, Scikit-Learn \cite{scikit-learn}, PyTorch \cite{NEURIPS2019_9015}, and OpenAI Gym \cite{brockman2016openai}.
We utilize the Open AI Gym framework to be compatible with several existing libraries.
This makes SOCKS comparable to several existing machine learning frameworks and promotes a more direct comparison with state-of-the-art machine learning and reinforcement learning algorithms.

%%%%%%%%%%%%%%%%%%%%%%%%%%%%%%%%%%%%%%%%%%%%%%%%%%

\subsection{Generating Samples}

The algorithms in SOCKS are data-driven, which means they rely upon a sample of system observations $\mathcal{S}$ as in Assumption \ref{assume: sample available}.
Thus, we have implemented several sampling functions in SOCKS in order to generate samples from a system via simulation when a priori data is unavailable.

The process for generating samples consists of defining a \emph{sample generator}, a function which generates a tuple contained within the sample $\mathcal{S}$. For example, to generate a sample $\mathcal{S} = \lbrace (x_{i}, u_{i}, y_{i}) \rbrace_{i=1}^{M}$ as in \eqref{eqn: sample}, we use the following code:
\begin{minted}[frame=lines]{Python}
# Setup code omitted.
state_sampler = random_sampler(env.state_space)
action_sampler = random_sampler(env.action_space)

@sample_generator
def sampler():
    state = next(state_sampler)
    action = next(action_sampler)

    env.state = state
    next_state, *_ = env.step(action)
    yield (state, action, next_state)

S = sample(sampler, sample_size=100)
\end{minted}
Here, the \mintinline{python}{sample_generator} function generates a single observation of the system, and the \mintinline{python}{sample} function computes a collection of $M = 100$ observations taken from the system.

SOCKS implements several commonly-used sample generators, including a one-step sample generator (shown above) and a trajectory generator, which generates samples of trajectories over multiple time steps of the form $\mathcal{S} = \lbrace (x_{i}, \upsilon_{i}, \xi_{i}) \rbrace_{i=1}^{M}$, where $x_{i} \in \mathcal{X}$ are the initial conditions, $\xi_{i}  = \lbrace x_{i}^{1}, \ldots, x_{i}^{N} \rbrace$ is the sequence of states at each time step over the time horizon $N \in \mathbb{N}$, and $\upsilon_{i}  = \lbrace u_{i}^{0}, \ldots, u_{i}^{N-1} \rbrace$ is a sequence of control actions taken from a policy $\pi$.

%%%%%%%%%%%%%%%%%%%%%%%%%%%%%%%%%%%%%%%%%%%%%%%%%%

\subsection{Stochastic Optimal Control}

SOCKS can be used to solve the stochastic optimal control problem in \eqref{eqn: optimal control problem}.
Given a sample $\mathcal{S}$ as in \eqref{eqn: sample}, we can approximate the integrals in \eqref{eqn: optimal control problem} using an estimate $\hat{m}(x, u)$ of the kernel distribution embedding $m(x, u)$, which can then be computed as Hilbert space inner products.

\begin{minted}[frame=lines]{python}
# Setup code omitted.
policy = KernelControlFwd(
    cost_fn=cost_fn,
    constraint_fn=constraint_fn,
)
policy.train(S, A)
\end{minted}
Here, \mintinline{python}{KernelControlFwd} class defines the algorithm, where we compute the optimal policy by minimizing the cost forward in time at each time step. The variables \mintinline{python}{S} and
\mintinline{python}{A} define a sample $\mathcal{S}$ taken from the system as in \eqref{eqn: sample} and a collection of admissible control actions in $\mathcal{U}$, respectively. The \mintinline{python}{cost_fn} and \mintinline{python}{constraint_fn} are user-defined functions which return a real value.
We can also solve the stochastic optimal control problem via dynamic programming (backward in time) by using \mintinline{python}{KernelControlBwd} in place of \mintinline{python}{KernelControlFwd}.

%%%%%%%%%%%%%%%%%%%%%%%%%%%%%%%%%%%%%%%%%%%%%%%%%%

\subsection{Stochastic Reachability}

We can solve the terminal-hitting time stochastic reachability problem using SOCKS. Given a sample $\mathcal{S}$ as in \eqref{eqn: sample}, we can compute an empirical estimate $\hat{m}(x, u)$ of $m(x, u)$, and (assuming the stochastic reachability value functions $V_{t}^{\pi}$, $t = 1, \ldots, N$, are in $\mathscr{H}$) we can approximate the stochastic reachability backward recursion by approximating the value function expectations in \eqref{eqn: stochastic reachability backward recursion} via Hilbert space inner products with the estimate $\hat{m}(x, u)$.
In other words, we define the approximate value functions $\bar{V}_{t}^{*} : \mathcal{X} \to [0, 1]$, $t = 0, \ldots, N$, and form an approximation of the stochastic reachability backward recursion, given by,
\begin{align}
    \bar{V}_{N}^{*}(x) &= V_{N}^{\pi}(x), \\
    \bar{V}_{t}^{*}(x) &= \sup_{\pi \in \Pi} \boldsymbol{1}_{\mathcal{K}}(x) \int_{\mathcal{U}} \langle \bar{V}_{t+1}^{\pi}, \hat{m}(x, v) \rangle_{\mathscr{H}} \pi(\mathrm{d} v \mid x),
\end{align}
where $V_{N}^{*}(x) = \boldsymbol{1}_{\mathcal{T}}(x)$, and $\mathcal{K}, \mathcal{T} \subseteq \mathscr{B}_{\mathcal{X}}$ are the safe set and target set, respectively.
Then as shown in \cite{thorpe2020model, thorpe2021approximate}, the solution to the approximate backward recursion, $\bar{V}_{0}^{*}(x_{0})$, is an approximation of the maximal stochastic reachability safety probabilities. See \cite{thorpe2020model} for more details.
\begin{minted}[frame=lines]{python}
# Setup code omitted.
alg = KernelMaximalSR(
    time_horizon=time_horizon,
    constraint_tube=constraint_tube,
    target_tube=target_tube,
    problem="THT",
)
alg.fit(S, A)
Pr = alg.predict(T)
\end{minted}
Here, \mintinline{python}{KernelMaximalSR} is the stochastic reachability algorithm class, \mintinline{python}{time_horizon} is the number of time steps, \mintinline{python}{S} is a sample taken i.i.d. from the system, \mintinline{python}{A} is a collection of admissible control actions, \mintinline{python}{T} is a collection of test (or evaluation) points, i.e. the points where we seek to evaluate the safety probabilities, \mintinline{python}{target_tube} and \mintinline{python}{constraint_tube} are sets defining $\mathcal{T}$ and $\mathcal{K}$, indexed by time, and \mintinline{python}{"THT"} specifies that we wish to solve the terminal-hitting time problem. In order to solve the first-hitting time problem, we simply replace \mintinline{python}{"THT"} with \mintinline{python}{"FHT"}.

This means we can evaluate the safety probabilities for a system under a given policy, and enables an analysis of the likelihood of respecting a set of pre-defined safety constraints given by $\mathcal{K}$, based in the same data-driven framework as the stochastic optimal controller synthesis in \eqref{eqn: approximate stochastic optimal control problem}. The primary difference between our approach and existing tools such as \cite{ray2019benchmarking}, is that our approach is not based in reinforcement learning, and does not guard against unsafe exploration of the state space (while collecting the sample $\mathcal{S}$), a well-known problem in safe RL \cite[cf.][]{garcia2015comprehensive}.

%%%%%%%%%%%%%%%%%%%%%%%%%%%%%%%%%%%%%%%%%%%%%%%%%%

\subsection{Forward Reachability}

We also implemented a forward reachable set estimator in SOCKS from \cite{thorpe2021learning}.
Let $\mathbb{P}_{N}$ be some distribution on the state space $\mathcal{X}$, and let $\mathcal{S} = \lbrace x_{i} \rbrace_{i=1}^{M}$ be a sample taken i.i.d. from $\mathbb{P}_{N}$.
The approximate forward reachable set classifier $\tilde{\mathscr{F}}$ is an estimate of the support of $\mathbb{P}$ and is computed as the solution to the following regularized least-squares problem:
\begin{equation}
    \tilde{F} = \arg \min_{f \in \mathscr{H}} \frac{1}{M} \sum_{i=1}^{M} \lVert k(x_{i}, \cdot) - f(x_{i}) \rVert_{\mathscr{H}}^{2} + \lambda \lVert f \rVert_{\mathscr{H}}^{2},
\end{equation}
where $\lambda > 0$ is the regularization parameter, and $k$ is a \emph{separating kernel} \cite[see][]{thorpe2021learning, de2014learning}. An RKHS $\mathscr{H}$ with kernel $k$ separates all subsets $C \subset \mathcal{X}$ if there exists a function $f \in \mathscr{H}$ such that for all $x \not\in C$, $f(x) \neq 0$, and $f(x^{\prime}) = 0$ for all $x^{\prime} \in C$.
The Abel kernel $k(x, x^{\prime}) = \exp(-\lVert x - x^{\prime} \rVert_{2}/\sigma)$, $\sigma > 0$, is a separating kernel, and is implemented in SOCKS. Note that a Gaussian RBF kernel is not a separating kernel, since constant functions are not included in a Gaussian RKHS \cite{steinwart2008support}.

The approximate forward reachable set is then given by
\begin{equation}
    \label{eqn: approximate forward reachable set}
    \tilde{\mathscr{F}} = \lbrace x \in \mathcal{X} \mid \tilde{F}(x) \geq 1 - \tau \rbrace,
\end{equation}
where $\tau$ is a threshold parameter, typically computed as $\tau = 1 - \min_{1 \leq i \leq M} \tilde{F}(x_{i})$, where $x_{i} \in \mathcal{S}$.

The approximate forward reachable set classifier can accommodate non-convex regions, and the approximation converges almost surely to the true classifier. However, the approximation obtained via the algorithm is not a guaranteed under- or over-approximation, though it does admit finite sample bounds \cite{de2014learning}. See \cite{thorpe2021learning} for more details.

%%%%%%%%%%%%%%%%%%%%%%%%%%%%%%%%%%%%%%%%%%%%%%%%%%

\subsection{Batch Processing}

The primary computational hurdle of the kernel-based approach in SOCKS is the matrix inverse term $W$ in \eqref{eqn: beta}, which is $\mathcal{O}(M^{3})$ in general, where $M$ is the sample size. Thus, the computation time scales polynomially as a function of the sample size.
In addition, the optimal control algorithms frequently involve storing very large, dense matrices that scale as a function of $M$, $T$ (the number of evaluation points) and $P$ (the number of admissible control actions).
The large matrix sizes can lead to memory storage issues on systems with low available memory. In order to account for this, SOCKS implements a batch processing variant for algorithms with large sample sizes, which computes the solution in smaller ``chunks''. This does not affect the result, but leads to longer computation times, since we must compute multiple matrix multiplications rather than a single multiplication with a large matrix.

%%%%%%%%%%%%%%%%%%%%%%%%%%%%%%%%%%%%%%%%%%%%%%%%%%

\subsection{Dynamical System Modeling}

OpenAI Gym currently implements several classical controls problems, including an inverted pendulum, a cart-pole system, and a ``mountain car''. These systems are contained within \emph{environments}, which encapsulate the dynamics, constraints, and cost for the problem.
Building on OpenAI Gym's standard framework, we have implemented a new type of learning environment, \mintinline{python}{DynamicalSystem}, which makes defining systems with dynamics easier.
In addition, we have implemented several benchmark systems in SOCKS that involve classical controls problems which are not included in OpenAI Gym, including:
\begin{enumerate*}[mode=unboxed, label=(\roman*)]
    \item
    a satellite rendezvous and docking problem based on Clohessy-Wiltshire-Hill (CWH) dynamics,
    \item
    an $n$-D stochastic chain of integrators,
    \item
    a nonholonomic vehicle,
    \item
    a point-mass system,
    \item
    a benchmark quadrotor example \cite{geretti2020arch},
    \item
    a planar quadrotor system,
    \item
    a translational oscillation with rotational actuation (TORA) system.
\end{enumerate*}
We plan to add additional benchmarks, since OpenAI Gym can also be used to simulate hybrid dynamics, partially observable systems, and more.

Simulating a \mintinline{python}{DynamicalSystem} can be done easily.
For example, we can evaluate the policy computed via the solution to the stochastic optimal control problem.
\begin{minted}[frame=lines]{Python}
# Setup code omitted.
env = NDIntegrator(2)
env.reset()
for t in range(time_horizon):
    action = policy(env.state)
    state, *_ = env.step(action)
\end{minted}
Here, \mintinline{python}{policy} is the result of the stochastic optimal control algorithm.
Behind the scenes, the simulation solves an initial value problem at each time step using Scipy's ODE solver. The input model is a zero-order hold.
This means we can simulate continuous dynamics in discrete time with different sampling times without redefining or parameterizing the dynamics.

%%%%%%%%%%%%%%%%%%%%%%%%%%%%%%%%%%%%%%%%%%%%%%%%%%
%%%%%%%%%%%%%%%%%%%%%%%%%%%%%%%%%%%%%%%%%%%%%%%%%%

\section{Numerical Experiments}
\label{section: numerical experiments}

All experiments were performed on an AWS cloud computing instance.
The toolbox and code to reproduce all results and analysis is available at \url{https://github.com/ajthor/socks}.

%%%%%%%%%%%%%%%%%%%%%%%%%%%%%%%%%%%%%%%%%%%%%%%%%%

\subsection{Nonholonomic Vehicle}

\begin{figure}
    \centering
    \includegraphics{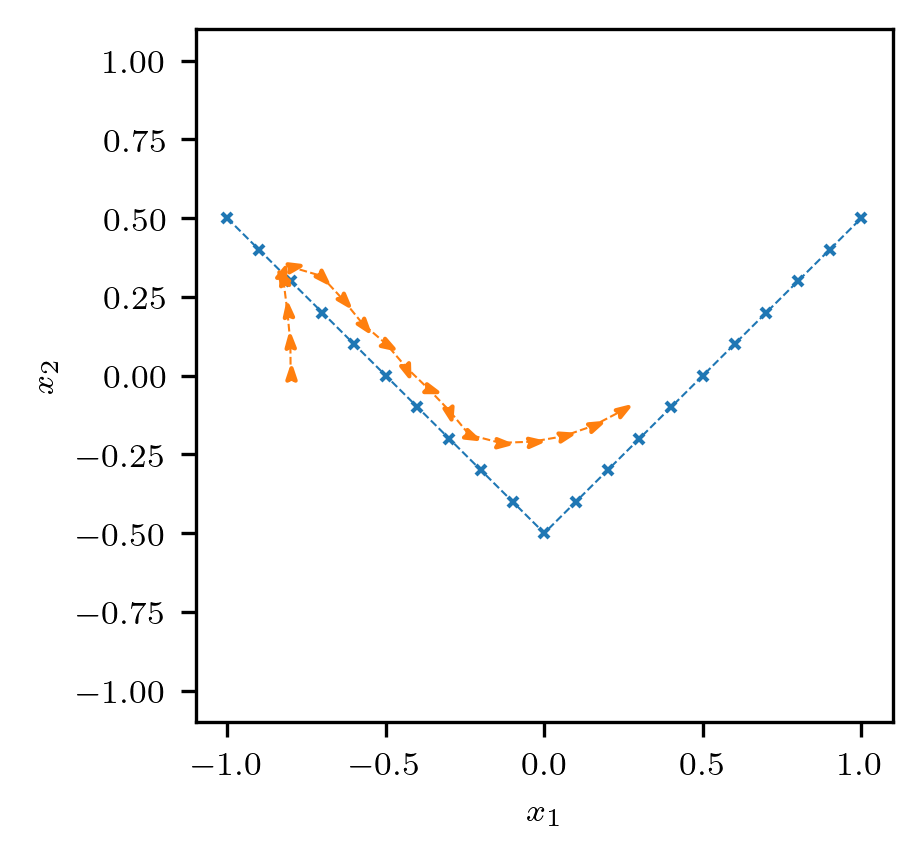}
    \includegraphics{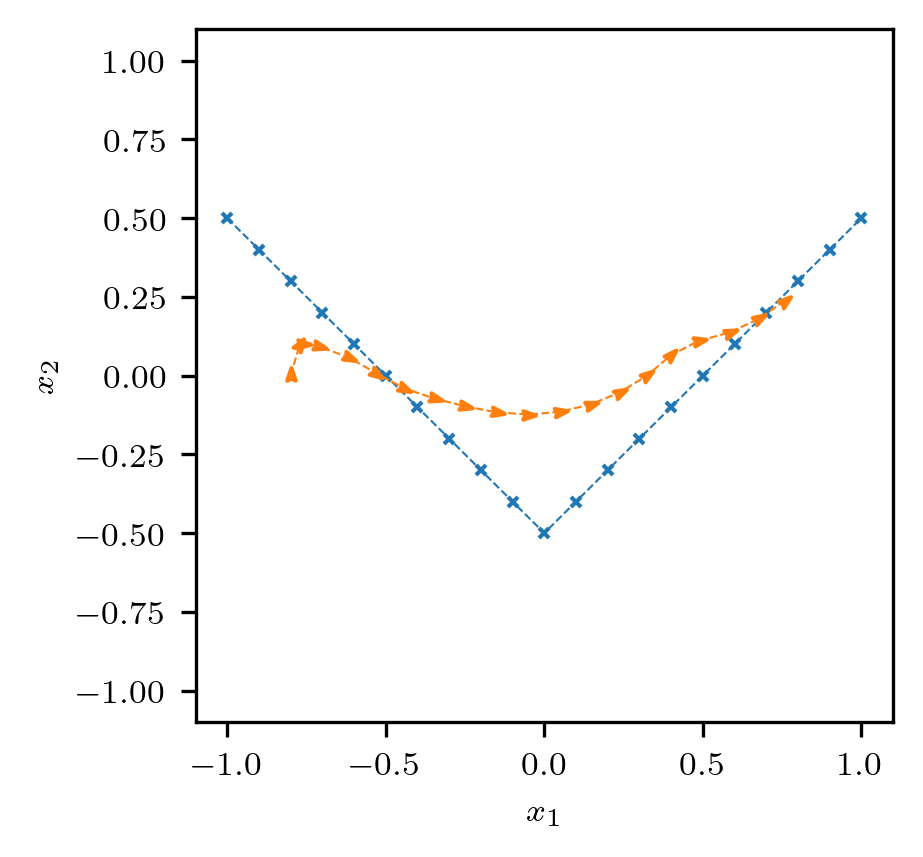}
    \caption{Nonholonomic vehicle trajectory using stochastic optimal control (top) and using dynamic programming (bottom) over a time horizon of $N = 20$. The dashed blue line shows the target trajectory, and the orange line shows the nonholonomic vehicle trajectory. Note that the dynamic programming trajectory better satisfies the terminal constraint.}
    \label{fig: nonholonomic}
\end{figure}

We consider a target tracking problem using nonholonomic vehicle dynamics, given by:
\begin{align}
    \dot{x}_{1} &= u_{1} \cos(x_{3}), &
    \dot{x}_{2} &= u_{1} \sin(x_{3}), &
    \dot{x}_{3} &= u_{2},
\end{align}
where $\mathcal{X} \subseteq \mathbb{R}^{3}$, $\mathcal{U} \subseteq \mathbb{R}^{2}$, and we constrain the input such that $u_{1} \in [0.1, 1]$, $u_{2} \in [-10, 10]$.
We then discretize the dynamics in time with sampling time $T_{s}$ and apply an affine stochastic disturbance $w \sim \mathcal{N}(0, \Sigma)$, $\Sigma = 0.01 I$.

The goal is to minimize the distance to an object moving along the v-shaped trajectory shown in blue in Figure \ref{fig: nonholonomic}.
We then collect a sample $\mathcal{S} = \lbrace(x_{i}, u_{i}, y_{i}) \rbrace_{i=1}^{M}$, of size $M = 2500$, and compute the optimal control actions using both the optimal control and dynamic programming algorithms in SOCKS with $\sigma = 2$. The results are shown in Figure \ref{fig: nonholonomic}, and computation time took approximately $0.204$ seconds for the optimal control algorithm and $12.003$ seconds for the dynamic programming algorithm.
We can see that the system more closely meets the terminal constraint using the dynamic programming algorithm, but the computation time increases dramatically, since we must compute a sequence of value functions.

%%%%%%%%%%%%%%%%%%%%%%%%%%%%%%%%%%%%%%%%%%%%%%%%%%

\subsection{Satellite Rendezvous and Docking}

\begin{figure}
    \centering
    \includegraphics[keepaspectratio]{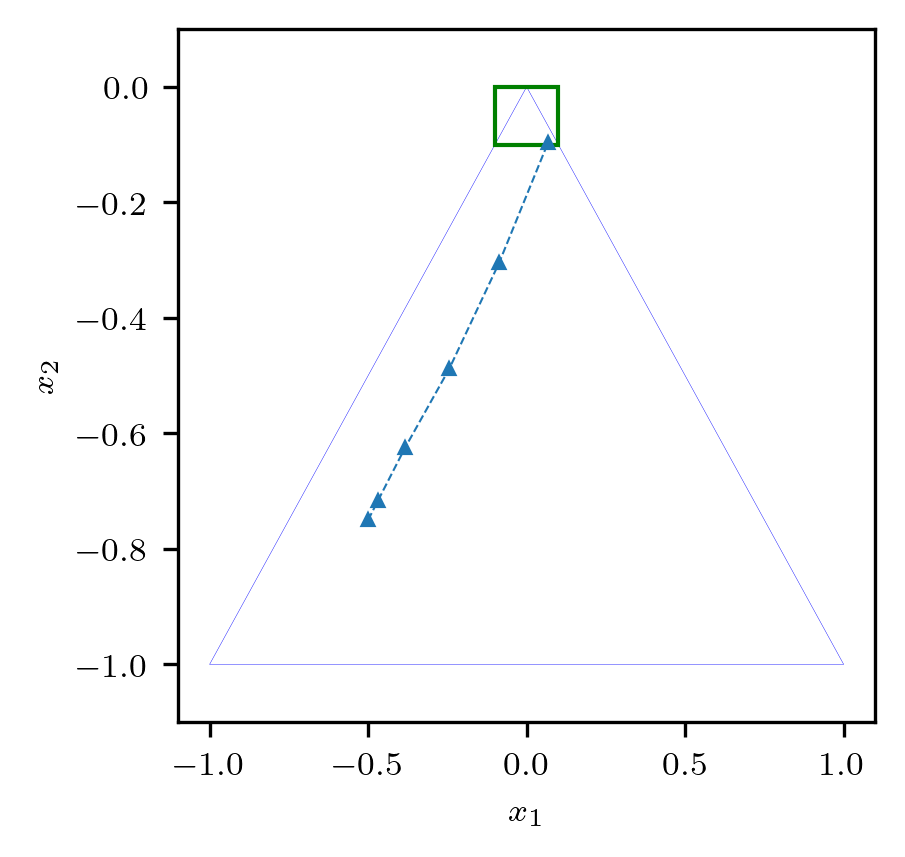}
    \caption{Optimal control for spacecraft rendezvous and docking problem with CWH dynamics. The goal is to reach a small region close to the origin (green square) while remaining within a line of sight cone (blue triangle). The trajectory of the system using the approximate stochastic optimal control algorithm is shown in blue.}
    \label{fig: cwh fwd}
\end{figure}

We consider an example of
spacecraft rendezvous and docking, in which one spacecraft must dock with another while remaining within a line of sight cone.
The Clohessy-Wiltshire-Hill dynamics are given by,
\begin{align}
    \label{eqn: cwh dynamics}
	\ddot{x} - 3\omega^{2} x - 2 \omega \dot{y} &= F_{x}/m_{d}, &
	\ddot{y} + 2 \omega \dot{x} &= F_{y}/m_{d},
\end{align}
with state $z = [x, y, \dot{x}, \dot{y}]^{\top} \in \mathcal{X} \subseteq \mathbb{R}^{4}$, input $u = [F_{x}, F_{y}]^{\top} \in \mathcal{U} \subseteq \mathbb{R}^{2}$, where $\mathcal{U} = [-0.1, 0.1] \times [-0.1, 0.1]$, and parameters $\omega$, $m_d$.
From \cite{lesser2013stochastic}, the dynamics in \eqref{eqn: cwh dynamics} can be written as a discrete-time LTI system $z_{t+1} = A z_{t} + B u_{t} + w_{t}$ with an additive Gaussian disturbance $w_{k} \sim \mathcal{N}(0, \Sigma)$, where $\Sigma = \textnormal{diag}([1 \times 10^{-4}, 1 \times 10^{-4}, 5 \times 10^{-8}, 5 \times 10^{-8}])$.

We apply the stochastic optimal control algorithm in SOCKS to the CWH system using the kernel bandwidth parameter $\sigma = 0.1$ and with a sample $\mathcal{S} = \lbrace(x_{i}, u_{i}, y_{i}) \rbrace_{i=1}^{M}$ of size $M = 2{,}500$ with
points $x_{i}$ sampled uniformly in the region $[-1.1, 1.1] \times [-1.1, 1.1] \times [-0.06, 0.06] \times [-0.06, 0.06]$, $u_{i} \in [-0.05, 0.05]^{2}$, and $y_{i} \sim Q(\cdot \mid x_{i}, u_{i})$.
The result is shown in Figure \ref{fig: cwh fwd}, and computation time was approximately $0.203$ seconds.

%%%%%%%%%%%%%%%%%%%%%%%%%%%%%%%%%%%%%%%%%%%%%%%%%%

\subsection{Double Integrator System}

\begin{figure}
    \centering
    \includegraphics[keepaspectratio]{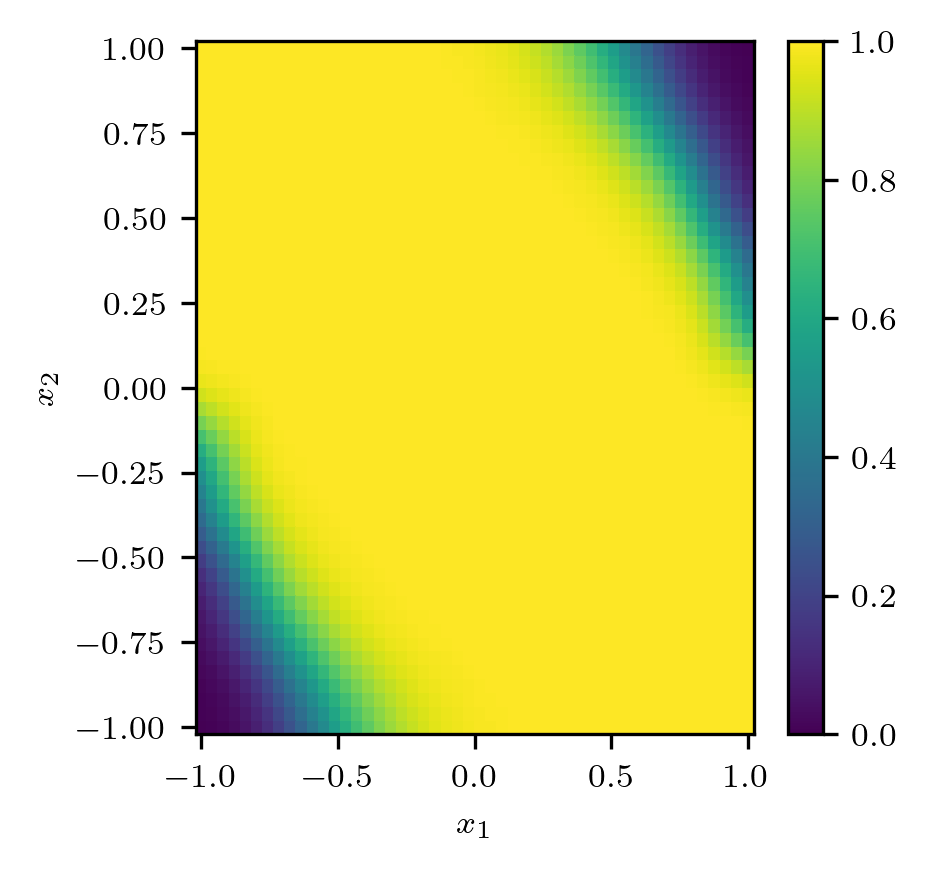}
    \caption{Stochastic reachability analysis of a double integrator system showing the maximal stochastic reachability safety probabilities with a safe set $\mathcal{K} = [-1, 1]^{2}$, target set $\mathcal{T} = [-0.5, 0.5]^{2}$, and time horizon $N = 16$ for the terminal-hitting time problem.}
    \label{fig: integrator sr}
\end{figure}

We consider a stochastic double integrator system in order to showcase the stochastic reachability analysis.
The dynamics of the system with sampling time $T_{s}$ are given by,
\begin{equation}
    \label{eqn: double integrator dynamics}
    x_{t+1} = \begin{bmatrix}
        1 & T_{s} \\
        0 & 1
    \end{bmatrix}
    x_{t} + \begin{bmatrix}
    T_{s}^{2}/2 \\
    T_{s}
    \end{bmatrix}
    u_{t} + w_{t},
\end{equation}
where $\mathcal{X} \subseteq \mathbb{R}^{2}$, $\mathcal{U} \subset \mathbb{R}$, and $w_{t} \sim \mathcal{N}(0, \Sigma)$, $\Sigma = 0.01 I$, is a random variable with a Gaussian distribution.
We collect a sample $\mathcal{S} = \lbrace (x_{i}, u_{i}, y_{i}) \rbrace_{i=1}^{M}$, $M = 2{,}500$, taken i.i.d. from $Q$, a representation of the dynamics in \eqref{eqn: double integrator dynamics} as a stochastic kernel, such that $x_{i}$ and $u_{i}$ are sampled uniformly from $\mathcal{X}$ and $\mathcal{U}$, respectively, with $x_{i}$ in the range $[-1.1, 1.1]^{2}$ and $u_{i}$ in the range $[-1, 1]$, and draw $y_{i}$ from $Q(\cdot \mid x_{i}, u_{i})$.
The safe set is defined as $\mathcal{K} = [-1, 1]^{2}$ and the target set is defined as $\mathcal{T} = [-0.5, 0.5]^{2}$.
We then computed the stochastic reachability safety probabilities for both the terminal-hitting time problem and the first-hitting time problem at $T = 10{,}000$ evaluation points using SOCKS and validated the result using Monte-Carlo.
The computation time was $\approx 3$ seconds for both problems.
The result is shown in Figure \ref{fig: integrator sr}.

%%%%%%%%%%%%%%%%%%%%%%%%%%%%%%%%%%%%%%%%%%%%%%%%%%

\subsection{Forward Reachability}

\begin{figure}
    \centering
    \includegraphics[width=3in,keepaspectratio]{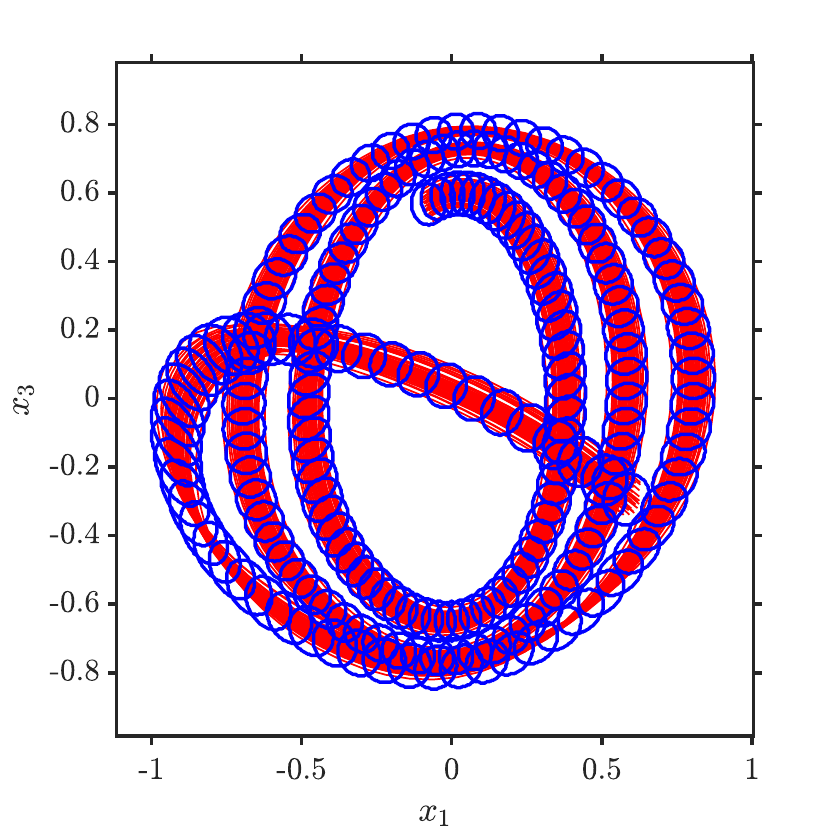}
    \caption{Approximate forward reachable set computed using the algorithm in \cite{thorpe2021learning} with sample size $M = 50$ and the Abel kernel with $\sigma = 0.1$.
    The solid blue lines indicate the estimated support boundary of the distribution at each time step and the red lines indicate the sampled trajectories.
    }
    \label{fig: forward reachability}
\end{figure}

We then demonstrate the forward reachable set algorithm in SOCKS for a translational oscillations by a rotational actuator (TORA) system \cite{thorpe2021learning}.
The dynamics of the system are given by,
\begin{align}
    \dot{x}_{1} &= x_{2},
    & \dot{x}_{2} &= -x_{1} + 0.1 \sin(x_{3}),
    & \dot{x}_{3} &= x_{4},
    & \dot{x}_{4} &= u,
\end{align}
where $\mathcal{X} \subseteq \mathbb{R}^{4}$, $\mathcal{U} \subset {R}$ and $u$ is a control input chosen by a neural network controller \cite{dutta2019sherlock}. We then discretize the dynamics in time and apply an affine stochastic disturbance $w \sim \mathcal{N}(0, \Sigma)$, $\Sigma = 0.01 I$.

We presume an initial distribution that is uniform over the region $[0.6, 0.7] \times [-0.7, -0.6] \times [-0.4, -0.3] \times [0.5, 0.6]$ and collect a sample $\mathcal{S}$ consisting of $M = 50$ simulated trajectories over a time horizon $N = 200$. Then, we apply the forward reachable set estimation algorithm and compute a classifier using \eqref{eqn: approximate forward reachable set} that estimates the support of the distribution at each time step.
The results are shown in Figure \ref{fig: forward reachability}, and the computation time was approximately $50.6$ seconds.

%%%%%%%%%%%%%%%%%%%%%%%%%%%%%%%%%%%%%%%%%%%%%%%%%%

\subsection{Scalability \& Computation Time}

\begin{figure*}
    \centering
    \includegraphics[keepaspectratio]{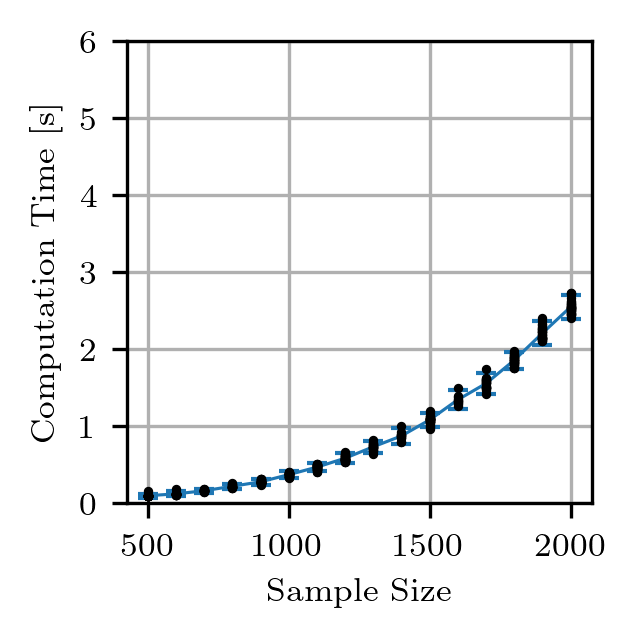}
    \includegraphics[keepaspectratio]{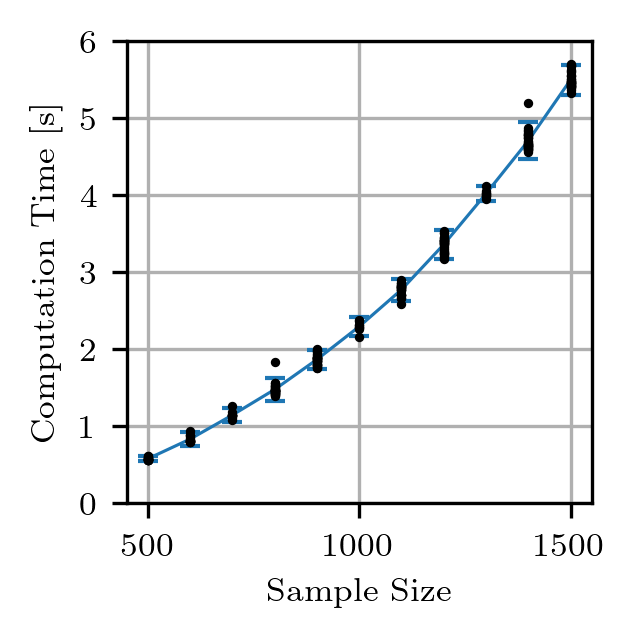}
    \includegraphics[keepaspectratio]{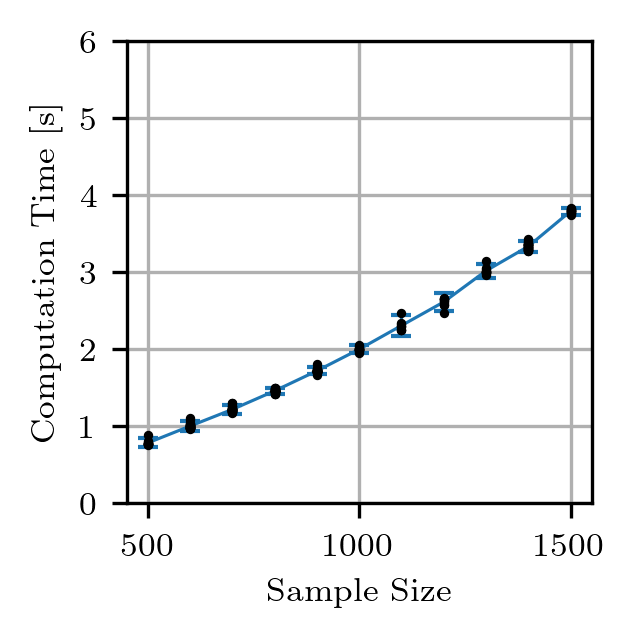}
    \caption{Computation time of the stochastic optimal control algorithm (left), the dynamic programming algorithm (center), and the maximal stochastic reachability algorithm (right) as a function of sample size $M$. The black dots indicate the computation time at each trial, and the blue bars indicate the 95\% confidence interval over 16 trials. The computation time scales polynomially as a function of sample size, and is generally $\mathcal{O}(M^{3})$.}
    \label{fig: computation as a function of sample size}
\end{figure*}

\begin{figure}
    \centering
    \includegraphics{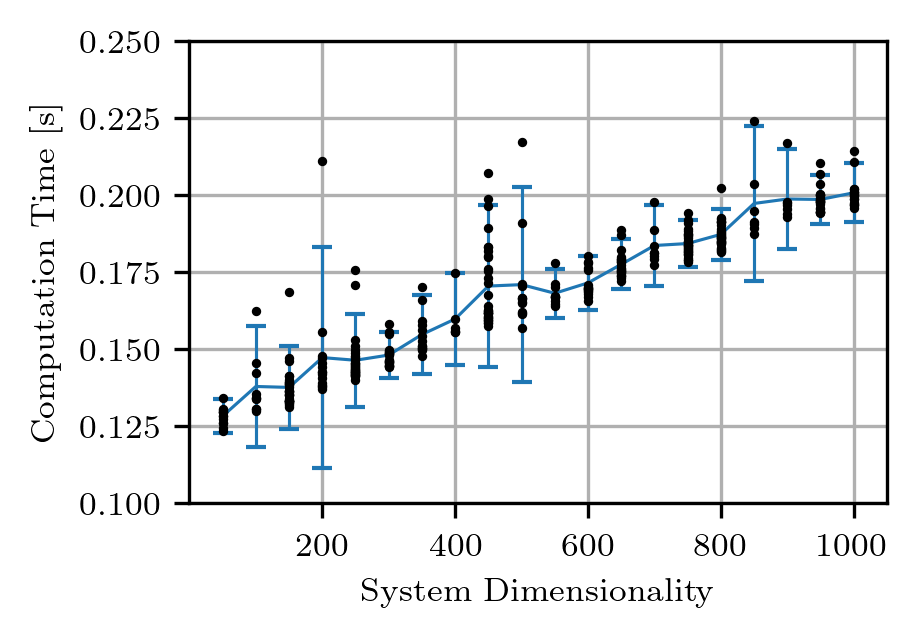}
    \caption{Computation time of the stochastic optimal control algorithm for an $n$-dimensional integrator system as a function of system dimensionality. The blue bars indicate the 95\% confidence interval over 16 trials. The computation time scales roughly linearly with the dimensionality of the system.}
    \label{fig: computation time as a function of dimensionality}
\end{figure}

We now present a brief discussion of the scalability and computational complexity of the algorithms.
As shown in \cite{thorpe2020model, thorpe2020sreachtools, thorpe2021stochastic}, the sample size $M$ used to compute the empirical distribution embedding $\hat{m}(x, u)$ presents the most significant computational burden, and is generally $\mathcal{O}(M^{3})$ due to the presence of the matrix inversion in \eqref{eqn: conditional distribution embedding estimate}.
We demonstrate this empirically for the algorithms presented in SOCKS in Figure \ref{fig: computation as a function of sample size}. We calculated the computation times for the algorithms over 16 runs for different values of $M$, and computed the statistical average and the 95\% confidence interval. The black dots indicate the empirically measured times, and the blue bars indicate the confidence interval for our algorithm. We can see that the computation times scale polynomially with the sample size $M$, as expected.
This can be prohibitive, since the quality of the kernel-based approximation improves as the sample size tends to infinity.
Nevertheless, several approximative speedup techniques (e.g. \cite{grunewalder2012conditional, rahimi2007random}) have been developed to alleviate the computational burden, and have been shown to reduce the computational complexity to $\mathcal{O}(M \log M)$. These techniques are not currently implemented in SOCKS, but we plan to include them as part of a future release.

As mentioned in \cite{thorpe2020model}, the complexity of the kernel-based algorithms scales roughly linearly with the dimensionality of the system.
This is primarily due to the fact that the system dimensionality only plays a role in the kernel evaluations, and does not significantly affect the computation of the empirical embedding $\hat{m}(x, u)$. We demonstrate this empirically for an $n$-dimensional stochastic chain of integrators system as in \eqref{eqn: double integrator dynamics} (see \cite{thorpe2020model}), where we choose a fixed sample size $M = 1000$ and vary the system dimensionality from $n = 50$ to $n = 1000$. The computation times are shown in Figure \ref{fig: computation time as a function of dimensionality}. We can see in Figure \ref{fig: computation time as a function of dimensionality} that as the system is dimensionality is increased, the computation time increases roughly linearly. However, as mentioned in \cite{thorpe2020model}, for high-dimensional systems, the sample size needed to fully characterize the dynamics and the uncertainty increases as the system dimensionality increases, which can be prohibitive for the reasons mentioned above.

%%%%%%%%%%%%%%%%%%%%%%%%%%%%%%%%%%%%%%%%%%%%%%%%%%
%%%%%%%%%%%%%%%%%%%%%%%%%%%%%%%%%%%%%%%%%%%%%%%%%%

\section{Conclusion \& Future Work}
\label{section: conclusion}

In this paper, we introduced SOCKS, a toolbox for approximate stochastic optimal control and approximate stochastic reachability based on a data-driven statistical learning technique known as kernel embeddings of distributions. We have demonstrated the capabilities of the toolbox on a simple stochastic chain of integrators example, a more realistic satellite rendezvous and docking example, a target tracking scenario using nonlinear nonholonomic vehicle dynamics, and on a forward reachable set estimation problem. The approaches used in SOCKS are scalable, computationally efficient, and model-free.

We plan to introduce additional kernel-based algorithms and features to SOCKS, such as the capability to handle neural network reachability analysis, trajectory optimization, and chance-constrained optimization.

\begin{acks}
    This material is based upon work supported by the \grantsponsor{1}{National Science Foundation}{}\ under NSF Grant Numbers \grantnum{1}{CNS-1836900} and \grantnum{1}{CMMI-2105631}. Any opinions, findings, and conclusions or recommendations expressed in this material are those of the authors and do not necessarily reflect the views of the National Science Foundation. The \grantsponsor{2}{NASA University Leadership initiative}{}\ (Grant \#\grantnum{2}{80NSSC20M0163}) provided funds to assist the authors with their research, but this article solely reflects the opinions and conclusions of its authors and not any NASA entity.
\end{acks}

%%%%%%%%%%%%%%%%%%%%%%%%%%%%%%%%%%%%%%%%%%%%%%%%%%
%%%%%%%%%%%%%%%%%%%%%%%%%%%%%%%%%%%%%%%%%%%%%%%%%%

\appendix

\section{Stochastic Optimal Control}
\label{appendix: stochastic optimal control}

In this section, we give an overview of the stochastic optimal control algorithm. Additional details are provided in \cite{thorpe2021stochastic}.
Given a sample $\mathcal{S}$ as in \eqref{eqn: sample}, taken i.i.d. from $Q$, we can compute an empirical estimate $\hat{m}(x, u)$ of the conditional distribution embedding $m(x, u)$.
By assumption \ref{assume: cost decomposed}, we assume that the objective and constraints can be decomposed as $f_{i}(x, u) = f_{i}^{x}(x) + f_{i}^{u}(u)$.
Assuming the objective function $f_{0}$ and constraints $f_{i}$, $i = 1, \ldots, p$, are elements of the RKHS $\mathscr{H}$, we can approximate the expectation with respect to $Q(\cdot \mid x, u)$ via the reproducing property of $k$ in $\mathscr{H}$,
\begin{align}
    & \int_{\mathcal{U}} \int_{\mathcal{X}} f_{i}(y, v) Q(\mathrm{d} y \mid x, v) \pi_{t}(\mathrm{d} v \mid x) \nonumber \\
    & \qquad = \int_{\mathcal{U}} \int_{\mathcal{X}} f_{i}^{x}(y) Q(\mathrm{d} y \mid x, v) + f_{i}^{u}(v) \pi_{t}(\mathrm{d} v \mid x) \\
    & \qquad \approx \int_{\mathcal{U}} \langle f_{i}^{x}, \hat{m}(x, v) \rangle_{\mathscr{H}} + f_{i}^{u}(v) \pi_{t}(\mathrm{d} v \mid x) \\
    \label{eqn: stochastic optimal control approximation}
    & \qquad = \int_{\mathcal{U}} \boldsymbol{f}_{i}^{x}{}^{\top} W \Psi k(x, \cdot) l(v, \cdot) + f_{i}^{u}(v) \pi_{t}(\mathrm{d} v \mid x),
\end{align}
for all $f_{i}$, $i = 0, 1, \ldots, p$.
Then, following \cite{thorpe2021stochastic}, we propose the following form for the approximation of the policy. Given a set $\mathcal{A} \subset \mathcal{U}$ of admissible control actions, $\mathcal{A} = \lbrace \tilde{u}_{j} \rbrace_{i=1}^{P}$, $P \in \mathbb{N}$, we have
\begin{equation}
    \label{eqn: policy approximation}
    \hat{p}_{t}(x) = \sum_{j=1}^{P} \gamma(x) l(\tilde{u}_{j}, \cdot).
\end{equation}
We can write \eqref{eqn: stochastic optimal control approximation} using the policy approximation in \eqref{eqn: policy approximation}, and obtain
\begin{align}
    & \int_{\mathcal{U}} \boldsymbol{f}_{i}^{x}{}^{\top} W \Psi k(x, \cdot) l(v, \cdot) + f_{i}^{u}(v) \pi_{t}(\mathrm{d} v \mid x) \nonumber \\
    & \qquad \approx
    \boldsymbol{f}_{i}^{x}{}^{\top} W \Psi \Upsilon^{\top} k(x, \cdot) \gamma(x) + \boldsymbol{f}_{i}^{u}{}^{\top} \gamma(x),
\end{align}
where $\Upsilon$ is a feature vector with elements $\Upsilon_{j} = l(\tilde{u}_{j}, \cdot)$.
We use this representation in the optimal control problem in order to approximate the objective and constraints. The following problem is approximately equivalent to the optimal control problem in \eqref{eqn: optimal control problem},
\begin{subequations}
    \begin{align}
        \min_{\gamma \in \mathbb{R}^{P}} \quad & \boldsymbol{f}_{0}^{x}{}^{\top} W \Psi \Upsilon^{\top} k(x, \cdot) \gamma(x) + \boldsymbol{f}_{0}^{u}{}^{\top} \gamma(x) \\
        \textnormal{s.t.} \quad & \boldsymbol{f}_{i}^{x}{}^{\top} W \Psi \Upsilon^{\top} k(x, \cdot) \gamma(x) + \boldsymbol{f}_{i}^{u}{}^{\top} \gamma(x) \leq 0, i = 1, \ldots, p
    \end{align}
\end{subequations}
Furthermore, as the number of observations $M$ in the sample $\mathcal{S}$ and the number of admissible control actions $P$ in $\mathcal{A}$ tends to infinity, the solution to the approximate problem converges in probability to the true solution \cite{thorpe2021stochastic}. See Appendix \ref{section: stability and convergence} for a more detailed discussion of convergence.
However, the optimization problem is unbounded below. In order to ensure feasibility, following \cite{thorpe2021stochastic}, we add an additional constraint such that $\gamma(x)$ lies in the probability simplex $\gamma(x) \in \lbrace a \mid \boldsymbol{1}^{\top} a = 1, 0 \preceq a \rbrace$, where $\boldsymbol{1}$ is a vector of all ones.
Thus, the approximate optimal control problem becomes
\begin{subequations}
    \label{eqn: approximate stochastic optimal control problem}
    \begin{align}
        \min_{\gamma \in \mathbb{R}^{P}} \quad & \boldsymbol{f}_{0}^{x}{}^{\top} W \Psi \Upsilon^{\top} k(x, \cdot) \gamma(x) + \boldsymbol{f}_{0}^{u}{}^{\top} \gamma(x) \\
        \textnormal{s.t.} \quad & \boldsymbol{f}_{i}^{x}{}^{\top} W \Psi \Upsilon^{\top} k(x, \cdot) \gamma(x) + \boldsymbol{f}_{i}^{u}{}^{\top} \gamma(x) \leq 0, i = 1, \ldots, p \\
        & \boldsymbol{1}^{\top} \gamma(x) = 1 \\
        & 0 \preceq \gamma(x)
    \end{align}
\end{subequations}
This is a linear program, and can be solved efficiently, e.g. via several commonly-used interior point or simplex algorithms \cite{boyd2004convex}.
Additionally, this representation can be used to solve a backward-in-time stochastic optimal control problem (dynamic programming).
See \cite{thorpe2021stochastic} for more details.

%%%%%%%%%%%%%%%%%%%%%%%%%%%%%%%%%%%%%%%%%%%%%%%%%%
%%%%%%%%%%%%%%%%%%%%%%%%%%%%%%%%%%%%%%%%%%%%%%%%%%

\section{Stability \& Convergence}
\label{section: stability and convergence}

We now seek to characterize the quality of the approximation and the conditions for its convergence.
The convergence properties of kernel distribution embeddings are well-studied in literature. See e.g. \cite{song2009hilbert, song2010nonparametric, grunewalder2012conditional, park2020measure} for more information.
However, the nuances of the different convergence results for kernel distribution embeddings means that the results do not always generalize well to all problems under different kernel choices.
For instance, the result in \cite[Theorem~6]{song2009hilbert} shows that the empirical estimate $\hat{m}(x, u)$ converges in probability to the true embedding $m(x, u)$ at a rate of $\mathcal{O}_{p}((M \lambda)^{-1/2} + \lambda^{1/2})$, where $\lambda$ is the regularization parameter in \eqref{eqn: regularized least squares} and $M$ is the sample size, but \cite{song2009hilbert} assumes that the RKHS is finite-dimensional, which does not hold for common kernel choices such as the Gaussian RBF kernel.
Thus, we present convergence results for the approximate stochastic optimal control problem based in the theory of algorithmic stability \cite{bousquet2002stability}.
Our result is close to the result presented in \cite{park2020measure}.

We first seek to characterize the convergence of the estimate $\hat{m}(x, u)$ in \eqref{eqn: conditional distribution embedding estimate} to its actual counterpart $m(x, u)$ in \eqref{eqn: conditional distribution embedding}.
For simplicity of notation, we define the operator $k_{x} : \mathcal{X} \to \mathbb{R}$ for all $x \in \mathcal{X}$ via $k_{x}(x^{\prime}) = k(x, x^{\prime})$.

Recall that the estimate $\hat{m}$ is in a vector-valued RKHS $\mathscr{Q}$ and is the solution to the regularized least-squares problem in \eqref{eqn: regularized least squares}.
Let $J : \mathscr{H} \times \mathscr{H} \to \mathbb{R}$ be a real-valued cost function, defined by
\begin{equation}
    J(k_{y}, \hat{m}(x, u)) \coloneqq \lVert k_{y} - \hat{m}(x, u) \rVert_{\mathscr{H}}^{2}.
\end{equation}
Let $\mathcal{Z} = \mathcal{X} \times \mathcal{U} \times \mathcal{X}$.
We define the \emph{loss function} $\upsilon : \mathscr{Q} \times \mathcal{Z} \to \mathbb{R}$, given by
\begin{equation}
    \upsilon(\hat{m}, (x, u, y)) = J(k_{y}, \hat{m}(x, u)).
\end{equation}

The \emph{risk}, denoted by $R(\hat{m})$, measures the expected loss (error) of the solution $\hat{m}$ to the regularized least-squares learning problem, and is defined as
\begin{equation}
    \label{eqn: risk}
    R(\hat{m}) \coloneqq \int_{\mathcal{X}} \upsilon(\hat{m}, (x, u, y)) Q(\mathrm{d} y \mid x, u).
\end{equation}
However, we cannot compute the risk directly since $Q$ is unknown by Assumption \ref{assume: stochastic kernel is unknown}. Thus, we seek to bound the risk by its empirical counterpart.
Given a sample $\mathcal{S} \in \mathcal{Z}^{M}$ as in \eqref{eqn: sample}, the \emph{empirical risk}, denoted by $R_{\mathcal{S}}(\hat{m})$, also known as the empirical error, measures the actual loss of the learning problem, and is defined as
\begin{align}
    \label{eqn: empirical risk}
    R_{\mathcal{S}}(\hat{m}) &\coloneqq \frac{1}{M} \sum_{i=1}^{M} \upsilon(\hat{m}, (x_{i}, u_{i}, y_{i})) + \lambda \lVert \hat{m} \rVert_{\mathscr{Q}}^{2} \\
    &= \frac{1}{M} \sum_{i=1}^{M} \lVert k_{y_{i}} - \hat{m}(x_{i}, u_{i}) \rVert_{\mathscr{H}}^{2} + \lambda \lVert \hat{m} \rVert_{\mathscr{Q}}^{2}.
\end{align}

We use $\hat{m}$ to denote the solution to the regularized least squares problem in \eqref{eqn: regularized least squares}, and let $\hat{m}^{\backslash i}$ denote the solution when a single observation is removed from $\mathcal{S}$ and let $\hat{m}^{i}$ denote the solution when the $i^{\rm th}$ observation is changed.
We use $\hat{m}^{\backslash i}$ and $\hat{m}^{i}$ in the following to assess the stability of the learning algorithm under minor changes to the sample $\mathcal{S}$ used to construct the estimate $\hat{m}$.

We present the following definition, modified from \cite{bousquet2002stability}, which allows us to characterize the stability of the learning algorithm with respect to the regularized least-squares problem in \eqref{eqn: regularized least squares}.

\begin{defn}[$\sigma$-admissible, {\cite[Definition~19]{bousquet2002stability}}]
    A loss function $\upsilon$ on $\mathscr{Q} \times \mathcal{Z}$ is $\sigma$-admissible with respect to $\mathscr{Q}$ if the associated cost function is convex with respect to its first argument and the following condition holds,
    \begin{equation}
        \lvert J(k_{y_{1}}, k_{y^{\prime}}) - J(k_{y_{2}}, k_{y^{\prime}}) \rvert \leq \sigma \lVert k_{y_{1}} - k_{y_{2}} \rVert_{\mathscr{H}}
    \end{equation}
    for all $k_{y^{\prime}} \in \mathscr{H}$ and $k_{y_{1}}, k_{y_{2}} \in \mathcal{D}$, where
    \begin{equation}
        \mathcal{D} = \lbrace k_{y} \mid \exists f \in \mathscr{Q}, \exists (x, u) \in \mathcal{X} \times \mathcal{U}, f(x, u) = k_{y} \rbrace
    \end{equation}
    is the domain of the first argument of $J$.
\end{defn}

We now seek to verify that the loss function $\upsilon$ pertaining to the regularized least-squares problem is $\sigma$-admissible with respect to $\mathscr{Q}$.
To this aim, we present the following proposition.

\begin{prop}
    The loss function given by
    \begin{equation}
        \upsilon(\hat{m}, (x, u, y)) = J(\hat{m}(x, u), k_{y}) = \lVert k_{y} - \hat{m}(x, u) \rVert_{\mathscr{H}}^{2}
    \end{equation}
    is $\sigma$-admissible with respect to $\mathscr{Q}$.
\end{prop}

The proof follows \cite[Lemma~20]{bousquet2002stability}, which shows that the loss function using a Hilbert space norm is $\sigma$-admissible, where $\sigma$ depends on the choice of kernel and the corresponding Hilbert space of functions $\mathscr{H}$.

We now present the following definition from \cite{bousquet2002stability}, modified to our particular formulation, which bounds the maximum difference in the loss function under minor variations to the sample $\mathcal{S}$.

\begin{defn}[uniform stability, {\cite[Definition~6]{bousquet2002stability}}]
    A learning algorithm has uniform stability $\alpha$ with respect to the loss function $\upsilon$ if the following holds:
    \begin{equation}
        \lVert \upsilon(\hat{m}, \cdot) - \upsilon(\hat{m}^{\backslash i}, \cdot) \rVert_{\infty} \leq \alpha,
    \end{equation}
    for all $\mathcal{S} \in \mathcal{Z}^{M}$ and $i = 1, \ldots, M$.
\end{defn}

In addition, an algorithm with uniform stability has the following property:
\begin{equation}
    \lvert \upsilon(\hat{m}, \cdot) - \upsilon(\hat{m}^{i}, \cdot) \rvert \leq 2 \alpha.
\end{equation}

As a consequence of the above definitions, \cite{bousquet2002stability} shows that the regularized least-squares problem in the scalar RKHS case has uniform stability.
We modify \cite[Theorem~22]{bousquet2002stability} to a vector-valued RKHS in the following theorem.

\begin{thm}
    \label{thm: uniform stability}
    Let $\mathscr{H}$ be an RKHS with kernel $k$ and $\mathscr{Q}$ be a vector-valued RKHS of functions on $\mathcal{X} \times \mathcal{U}$ mapping to $\mathscr{H}$. Let $k$ be bounded by $\rho < \infty$, and let $\upsilon$ be a $\sigma$-admissible loss function with respect to $\mathscr{Q}$. Then the learning algorithm given by
    \begin{equation}
        \hat{m} = \arg \min_{f \in \mathscr{Q}} \frac{1}{M} \sum_{i=1}^{M} \upsilon(f, (y_{i}, x_{i}, u_{i})) + \lambda \lVert f \rVert_{\mathscr{Q}}^{2},
    \end{equation}
    has uniform stability $\alpha$ with respect to $\upsilon$ with
    \begin{equation}
        \label{eqn: uniform stability bound}
        \alpha \leq \frac{\sigma^{2} \rho^{2}}{2 \lambda M}.
    \end{equation}
\end{thm}

We can ensure boundedness of the kernel of $\mathscr{Q}$ using the principle of uniform boundedness (also known as the Banach-Steinhaus theorem), since the kernel $k$ is bounded by $\rho$.
Then the proof follows directly from \cite[Theorem~22]{bousquet2002stability}.

We use this result to show that the regularized least-squares problem in \eqref{eqn: regularized least squares} has uniform stability with respect to $\upsilon$.

\begin{thm}[{\cite[Theorem~12]{bousquet2002stability}}]
    \label{thm: risk bound}
    Let $A$ be an algorithm with uniform stability $\alpha$ with respect to a loss function $\upsilon$ such that $0 \leq \upsilon(\hat{m}, (x, u, y)) \leq B$, for all $(x, u, y) \in \mathcal{Z}$ and all sets $\mathcal{S}$. Then for any $M \geq 1$ and any $\delta \in (0, 1)$ the following bounds hold with probability $1 - \delta$ of the random draw of the sample $\mathcal{S}$:
    \begin{equation}
        R(\hat{m}) \leq R_{\mathcal{S}}(\hat{m}) + 2 \alpha + (4 M \alpha + B) \sqrt{\frac{\log(1/ \delta)}{2 M}}.
    \end{equation}
\end{thm}

Thus, using Theorem \ref{thm: risk bound} with $\alpha$ given by \eqref{eqn: uniform stability bound} from Theorem \ref{thm: uniform stability}, we have that for any $M \geq 1$ and any $\delta \in (0, 1)$, with probability $1 - \delta$, the risk $R$ is bounded by:
\begin{equation}
    R(\hat{m}) \leq R_{\mathcal{S}}(\hat{m}) + \frac{\sigma^{2} \rho^{2}}{\lambda M} + \biggl(\frac{2 \sigma^{2} \rho^{2}}{\lambda} + \rho \biggr) \sqrt{\frac{\log(1/\delta)}{2 M}},
\end{equation}
which shows that as the sample size $M$ increases, the empirical embedding $\hat{m}(x, u)$ in \eqref{eqn: conditional distribution embedding estimate} converges in probability to the true embedding $m(x, u)$ in \eqref{eqn: conditional distribution embedding}.
Thus, the approximation of the expectations in \eqref{eqn: approximate stochastic optimal control problem} converge in probability to the true expectations, and the approximate optimization problems computed using the estimate $\hat{m}(x, u)$ converge to the true optimization problems as $M$ increases.

Similarly, the approximate policy $\hat{p}(x)$ in \eqref{eqn: policy approximation} has the form of an empirical conditional distribution embedding as in \eqref{eqn: conditional distribution embedding estimate}, which suggests that the approximate policy (and consequently the approximately optimal control action) obtained via \eqref{eqn: approximate stochastic optimal control problem} also converges in probability as the number of admissible control actions $P$ in \eqref{eqn: policy approximation} is increased.

%%%%%%%%%%%%%%%%%%%%%%%%%%%%%%%%%%%%%%%%%%%%%%%%%%
%%%%%%%%%%%%%%%%%%%%%%%%%%%%%%%%%%%%%%%%%%%%%%%%%%

\bibliographystyle{ACM-Reference-Format}
\bibliography{bibliography.bib}


\begin{thebibliography}{57}

%%% the Web address.

\ifx \showCODEN    \undefined \def \showCODEN     #1{\unskip}     \fi
\ifx \showDOI      \undefined \def \showDOI       #1{#1}\fi
\ifx \showISBNx    \undefined \def \showISBNx     #1{\unskip}     \fi
\ifx \showISBNxiii \undefined \def \showISBNxiii  #1{\unskip}     \fi
\ifx \showISSN     \undefined \def \showISSN      #1{\unskip}     \fi
\ifx \showLCCN     \undefined \def \showLCCN      #1{\unskip}     \fi
\ifx \shownote     \undefined \def \shownote      #1{#1}          \fi
\ifx \showarticletitle \undefined \def \showarticletitle #1{#1}   \fi
\ifx \showURL      \undefined \def \showURL       {\relax}        \fi
% invisible to TeX
\providecommand\bibfield[2]{#2}
\providecommand\bibinfo[2]{#2}
\providecommand\natexlab[1]{#1}
\providecommand\showeprint[2][]{arXiv:#2}

\bibitem[\protect\citeauthoryear{Abadi, Agarwal, Barham, Brevdo, Chen, Citro,
  Corrado, Davis, Dean, Devin, Ghemawat, Goodfellow, Harp, Irving, Isard, Jia,
  Jozefowicz, Kaiser, Kudlur, Levenberg, Man\'{e}, Monga, Moore, Murray, Olah,
  Schuster, Shlens, Steiner, Sutskever, Talwar, Tucker, Vanhoucke, Vasudevan,
  Vi\'{e}gas, Vinyals, Warden, Wattenberg, Wicke, Yu, and Zheng}{Abadi
  et~al\mbox{.}}{2015}]%
        {tensorflow2015-whitepaper}
\bibfield{author}{\bibinfo{person}{Mart\'{\i}n Abadi}, \bibinfo{person}{Ashish
  Agarwal}, \bibinfo{person}{Paul Barham}, \bibinfo{person}{Eugene Brevdo},
  \bibinfo{person}{Zhifeng Chen}, \bibinfo{person}{Craig Citro},
  \bibinfo{person}{Greg~S. Corrado}, \bibinfo{person}{Andy Davis},
  \bibinfo{person}{Jeffrey Dean}, \bibinfo{person}{Matthieu Devin},
  \bibinfo{person}{Sanjay Ghemawat}, \bibinfo{person}{Ian Goodfellow},
  \bibinfo{person}{Andrew Harp}, \bibinfo{person}{Geoffrey Irving},
  \bibinfo{person}{Michael Isard}, \bibinfo{person}{Yangqing Jia},
  \bibinfo{person}{Rafal Jozefowicz}, \bibinfo{person}{Lukasz Kaiser},
  \bibinfo{person}{Manjunath Kudlur}, \bibinfo{person}{Josh Levenberg},
  \bibinfo{person}{Dandelion Man\'{e}}, \bibinfo{person}{Rajat Monga},
  \bibinfo{person}{Sherry Moore}, \bibinfo{person}{Derek Murray},
  \bibinfo{person}{Chris Olah}, \bibinfo{person}{Mike Schuster},
  \bibinfo{person}{Jonathon Shlens}, \bibinfo{person}{Benoit Steiner},
  \bibinfo{person}{Ilya Sutskever}, \bibinfo{person}{Kunal Talwar},
  \bibinfo{person}{Paul Tucker}, \bibinfo{person}{Vincent Vanhoucke},
  \bibinfo{person}{Vijay Vasudevan}, \bibinfo{person}{Fernanda Vi\'{e}gas},
  \bibinfo{person}{Oriol Vinyals}, \bibinfo{person}{Pete Warden},
  \bibinfo{person}{Martin Wattenberg}, \bibinfo{person}{Martin Wicke},
  \bibinfo{person}{Yuan Yu}, {and} \bibinfo{person}{Xiaoqiang Zheng}.}
  \bibinfo{year}{2015}\natexlab{}.
\newblock \bibinfo{title}{{TensorFlow}: Large-Scale Machine Learning on
  Heterogeneous Systems}.
\newblock
\newblock
\urldef\tempurl%
\url{https://www.tensorflow.org/}
\showURL{%
\tempurl}
\newblock
\shownote{Software available from tensorflow.org.}


\bibitem[\protect\citeauthoryear{Abate, Blom, Cauchi, Degiorgio, Fr{\"a}nzle,
  Hahn, Haesaert, Ma, Oishi, Pilch, et~al\mbox{.}}{Abate et~al\mbox{.}}{2019}]%
        {abate2019arch}
\bibfield{author}{\bibinfo{person}{Alessandro Abate}, \bibinfo{person}{Henk
  Blom}, \bibinfo{person}{Nathalie Cauchi}, \bibinfo{person}{Kurt Degiorgio},
  \bibinfo{person}{Martin Fr{\"a}nzle}, \bibinfo{person}{Ernst~Moritz Hahn},
  \bibinfo{person}{Sofie Haesaert}, \bibinfo{person}{Hao Ma},
  \bibinfo{person}{Meeko Oishi}, \bibinfo{person}{Carina Pilch},
  {et~al\mbox{.}}} \bibinfo{year}{2019}\natexlab{}.
\newblock \showarticletitle{{ARCH-COMP}19 category report: Stochastic
  modelling}.
\newblock \bibinfo{journal}{\emph{{EPiC} Series in Computing}}
  \bibinfo{volume}{61} (\bibinfo{year}{2019}), \bibinfo{pages}{62--102}.
\newblock


\bibitem[\protect\citeauthoryear{Abate, Blom, Cauchi, Delicaris, Hartmanns,
  Khaled, Lavaei, Pilch, Remke, Schupp, et~al\mbox{.}}{Abate
  et~al\mbox{.}}{2020}]%
        {abate2020arch}
\bibfield{author}{\bibinfo{person}{Alessandro Abate}, \bibinfo{person}{Henk
  Blom}, \bibinfo{person}{Nathalie Cauchi}, \bibinfo{person}{Joanna Delicaris},
  \bibinfo{person}{Arnd Hartmanns}, \bibinfo{person}{Mahmoud Khaled},
  \bibinfo{person}{Abolfazl Lavaei}, \bibinfo{person}{Carina Pilch},
  \bibinfo{person}{Anne Remke}, \bibinfo{person}{Stefan Schupp},
  {et~al\mbox{.}}} \bibinfo{year}{2020}\natexlab{}.
\newblock \showarticletitle{{ARCH-COMP}20 Category Report: Stochastic Models}.
\newblock \bibinfo{journal}{\emph{{EPiC} Series in Computing}}
  \bibinfo{volume}{74} (\bibinfo{year}{2020}), \bibinfo{pages}{76--106}.
\newblock


\bibitem[\protect\citeauthoryear{Abate, Blom, Cauchi, Haesaert, Hartmanns,
  Lesser, Oishi, Sivaramakrishnan, and Soudjani}{Abate et~al\mbox{.}}{2018}]%
        {abate2018arch}
\bibfield{author}{\bibinfo{person}{Alessandro Abate}, \bibinfo{person}{HAP
  Blom}, \bibinfo{person}{Nathalie Cauchi}, \bibinfo{person}{Sofie Haesaert},
  \bibinfo{person}{Arnd Hartmanns}, \bibinfo{person}{Kendra Lesser},
  \bibinfo{person}{Meeko Oishi}, \bibinfo{person}{Vignesh Sivaramakrishnan},
  {and} \bibinfo{person}{Sadegh Soudjani}.} \bibinfo{year}{2018}\natexlab{}.
\newblock \showarticletitle{{ARCH-COMP}18 Category Report: Stochastic
  Modelling}.
\newblock \bibinfo{journal}{\emph{{EPiC} Series in Computing}}
  \bibinfo{volume}{54} (\bibinfo{year}{2018}).
\newblock


\bibitem[\protect\citeauthoryear{Abate, Prandini, Lygeros, and Sastry}{Abate
  et~al\mbox{.}}{2008}]%
        {abate2008probabilistic}
\bibfield{author}{\bibinfo{person}{Alessandro Abate}, \bibinfo{person}{Maria
  Prandini}, \bibinfo{person}{John Lygeros}, {and} \bibinfo{person}{Shankar
  Sastry}.} \bibinfo{year}{2008}\natexlab{}.
\newblock \showarticletitle{Probabilistic reachability and safety for
  controlled discrete time stochastic hybrid systems}.
\newblock \bibinfo{journal}{\emph{Automatica}} \bibinfo{volume}{44},
  \bibinfo{number}{11} (\bibinfo{year}{2008}), \bibinfo{pages}{2724--2734}.
\newblock


\bibitem[\protect\citeauthoryear{Aronszajn}{Aronszajn}{1950}]%
        {aronszajn1950theory}
\bibfield{author}{\bibinfo{person}{Nachman Aronszajn}.}
  \bibinfo{year}{1950}\natexlab{}.
\newblock \showarticletitle{Theory of reproducing kernels}.
\newblock \bibinfo{journal}{\emph{Transactions of the American mathematical
  society}} \bibinfo{volume}{68}, \bibinfo{number}{3} (\bibinfo{year}{1950}),
  \bibinfo{pages}{337--404}.
\newblock


\bibitem[\protect\citeauthoryear{Bertsekas and Shreve}{Bertsekas and
  Shreve}{1978}]%
        {bertsekas1978stochastic}
\bibfield{author}{\bibinfo{person}{Dimitri~P Bertsekas} {and}
  \bibinfo{person}{Steven~E Shreve}.} \bibinfo{year}{1978}\natexlab{}.
\newblock \bibinfo{booktitle}{\emph{Stochastic optimal control: the discrete
  time case}}.
\newblock \bibinfo{publisher}{Elsevier}.
\newblock


\bibitem[\protect\citeauthoryear{Bousquet and Elisseeff}{Bousquet and
  Elisseeff}{2002}]%
        {bousquet2002stability}
\bibfield{author}{\bibinfo{person}{Olivier Bousquet} {and}
  \bibinfo{person}{Andr{\'e} Elisseeff}.} \bibinfo{year}{2002}\natexlab{}.
\newblock \showarticletitle{Stability and generalization}.
\newblock \bibinfo{journal}{\emph{The Journal of Machine Learning Research}}
  \bibinfo{volume}{2} (\bibinfo{year}{2002}), \bibinfo{pages}{499--526}.
\newblock


\bibitem[\protect\citeauthoryear{Boyd, Boyd, and Vandenberghe}{Boyd
  et~al\mbox{.}}{2004}]%
        {boyd2004convex}
\bibfield{author}{\bibinfo{person}{Stephen Boyd}, \bibinfo{person}{Stephen~P
  Boyd}, {and} \bibinfo{person}{Lieven Vandenberghe}.}
  \bibinfo{year}{2004}\natexlab{}.
\newblock \bibinfo{booktitle}{\emph{Convex optimization}}.
\newblock \bibinfo{publisher}{Cambridge university press}.
\newblock


\bibitem[\protect\citeauthoryear{Brockman, Cheung, Pettersson, Schneider,
  Schulman, Tang, and Zaremba}{Brockman et~al\mbox{.}}{2016}]%
        {brockman2016openai}
\bibfield{author}{\bibinfo{person}{Greg Brockman}, \bibinfo{person}{Vicki
  Cheung}, \bibinfo{person}{Ludwig Pettersson}, \bibinfo{person}{Jonas
  Schneider}, \bibinfo{person}{John Schulman}, \bibinfo{person}{Jie Tang},
  {and} \bibinfo{person}{Wojciech Zaremba}.} \bibinfo{year}{2016}\natexlab{}.
\newblock \showarticletitle{Openai gym}.
\newblock \bibinfo{journal}{\emph{arXiv preprint arXiv:1606.01540}}
  (\bibinfo{year}{2016}).
\newblock


\bibitem[\protect\citeauthoryear{Cauchi and Abate}{Cauchi and Abate}{2019}]%
        {cauchi2019stochy}
\bibfield{author}{\bibinfo{person}{Nathalie Cauchi} {and}
  \bibinfo{person}{Alessandro Abate}.} \bibinfo{year}{2019}\natexlab{}.
\newblock \showarticletitle{StocHy - Automated Verification and Synthesis of
  Stochastic Processes: Poster Abstract}. In
  \bibinfo{booktitle}{\emph{Proceedings of the 22nd ACM International
  Conference on Hybrid Systems: Computation and Control}}
  \emph{(\bibinfo{series}{HSCC '19})}. \bibinfo{publisher}{Association for
  Computing Machinery}, \bibinfo{address}{New York, NY, USA},
  \bibinfo{pages}{258–259}.
\newblock
\showISBNx{9781450362825}


\bibitem[\protect\citeauthoryear{Chollet et~al\mbox{.}}{Chollet
  et~al\mbox{.}}{2015}]%
        {chollet2015keras}
\bibfield{author}{\bibinfo{person}{Fran\c{c}ois Chollet} {et~al\mbox{.}}}
  \bibinfo{year}{2015}\natexlab{}.
\newblock \bibinfo{title}{Keras}.
\newblock \bibinfo{howpublished}{\url{https://keras.io}}.
\newblock


\bibitem[\protect\citeauthoryear{{\c{C}}{\i}nlar}{{\c{C}}{\i}nlar}{2011}]%
        {cinlar2011probability}
\bibfield{author}{\bibinfo{person}{Erhan {\c{C}}{\i}nlar}.}
  \bibinfo{year}{2011}\natexlab{}.
\newblock \bibinfo{booktitle}{\emph{Probability and Stochastics}}.
  Vol.~\bibinfo{volume}{261}.
\newblock \bibinfo{publisher}{Springer Science \& Business Media}.
\newblock


\bibitem[\protect\citeauthoryear{De~Vito, Rosasco, and Toigo}{De~Vito
  et~al\mbox{.}}{2014}]%
        {de2014learning}
\bibfield{author}{\bibinfo{person}{Ernesto De~Vito}, \bibinfo{person}{Lorenzo
  Rosasco}, {and} \bibinfo{person}{Alessandro Toigo}.}
  \bibinfo{year}{2014}\natexlab{}.
\newblock \showarticletitle{Learning sets with separating kernels}.
\newblock \bibinfo{journal}{\emph{Applied and Computational Harmonic Analysis}}
  \bibinfo{volume}{37}, \bibinfo{number}{2} (\bibinfo{year}{2014}),
  \bibinfo{pages}{185--217}.
\newblock


\bibitem[\protect\citeauthoryear{Dehnert, Junges, Katoen, and Volk}{Dehnert
  et~al\mbox{.}}{2017}]%
        {storm}
\bibfield{author}{\bibinfo{person}{Christian Dehnert},
  \bibinfo{person}{Sebastian Junges}, \bibinfo{person}{Joost-Pieter Katoen},
  {and} \bibinfo{person}{Matthias Volk}.} \bibinfo{year}{2017}\natexlab{}.
\newblock \showarticletitle{A Storm is Coming: A Modern Probabilistic Model
  Checker}. In \bibinfo{booktitle}{\emph{Computer Aided Verification}},
  \bibfield{editor}{\bibinfo{person}{Rupak Majumdar} {and}
  \bibinfo{person}{Viktor Kun{\v{c}}ak}} (Eds.). \bibinfo{publisher}{Springer
  International Publishing}, \bibinfo{address}{Cham},
  \bibinfo{pages}{592--600}.
\newblock


\bibitem[\protect\citeauthoryear{Deisenroth, Rasmussen, and Peters}{Deisenroth
  et~al\mbox{.}}{2009}]%
        {deisenroth2009gaussian}
\bibfield{author}{\bibinfo{person}{Marc~Peter Deisenroth},
  \bibinfo{person}{Carl~Edward Rasmussen}, {and} \bibinfo{person}{Jan Peters}.}
  \bibinfo{year}{2009}\natexlab{}.
\newblock \showarticletitle{Gaussian process dynamic programming}.
\newblock \bibinfo{journal}{\emph{Neurocomputing}} \bibinfo{volume}{72},
  \bibinfo{number}{7-9} (\bibinfo{year}{2009}), \bibinfo{pages}{1508--1524}.
\newblock


\bibitem[\protect\citeauthoryear{Djeumou and Topcu}{Djeumou and Topcu}{2021}]%
        {djeumou2021learning}
\bibfield{author}{\bibinfo{person}{Franck Djeumou} {and} \bibinfo{person}{Ufuk
  Topcu}.} \bibinfo{year}{2021}\natexlab{}.
\newblock \showarticletitle{Learning to Reach, Swim, Walk and Fly in One Trial:
  Data-Driven Control with Scarce Data and Side Information}.
\newblock \bibinfo{journal}{\emph{arXiv preprint arXiv:2106.10533}}
  (\bibinfo{year}{2021}).
\newblock


\bibitem[\protect\citeauthoryear{Djeumou, Zutshi, and Topcu}{Djeumou
  et~al\mbox{.}}{2021}]%
        {djeumou2021fly}
\bibfield{author}{\bibinfo{person}{Franck Djeumou}, \bibinfo{person}{Aditya
  Zutshi}, {and} \bibinfo{person}{Ufuk Topcu}.}
  \bibinfo{year}{2021}\natexlab{}.
\newblock \showarticletitle{On-the-fly, data-driven reachability analysis and
  control of unknown systems: an F-16 aircraft case study}. In
  \bibinfo{booktitle}{\emph{Proceedings of the 24th International Conference on
  Hybrid Systems: Computation and Control}}. \bibinfo{pages}{1--2}.
\newblock


\bibitem[\protect\citeauthoryear{Dutta, Chen, Jha, Sankaranarayanan, and
  Tiwari}{Dutta et~al\mbox{.}}{2019}]%
        {dutta2019sherlock}
\bibfield{author}{\bibinfo{person}{Souradeep Dutta}, \bibinfo{person}{Xin
  Chen}, \bibinfo{person}{Susmit Jha}, \bibinfo{person}{Sriram
  Sankaranarayanan}, {and} \bibinfo{person}{Ashish Tiwari}.}
  \bibinfo{year}{2019}\natexlab{}.
\newblock \showarticletitle{Sherlock-A tool for verification of neural network
  feedback systems}. In \bibinfo{booktitle}{\emph{International Conference on
  Hybrid Systems: Computation and Control}}. \bibinfo{pages}{262--263}.
\newblock


\bibitem[\protect\citeauthoryear{Garc{\i}a and Fern{\'a}ndez}{Garc{\i}a and
  Fern{\'a}ndez}{2015}]%
        {garcia2015comprehensive}
\bibfield{author}{\bibinfo{person}{Javier Garc{\i}a} {and}
  \bibinfo{person}{Fernando Fern{\'a}ndez}.} \bibinfo{year}{2015}\natexlab{}.
\newblock \showarticletitle{A comprehensive survey on safe reinforcement
  learning}.
\newblock \bibinfo{journal}{\emph{Journal of Machine Learning Research}}
  \bibinfo{volume}{16}, \bibinfo{number}{1} (\bibinfo{year}{2015}),
  \bibinfo{pages}{1437--1480}.
\newblock


\bibitem[\protect\citeauthoryear{Geretti, Sandretto, Althoff, Benet, Chapoutot,
  Chen, Collins, Forets, Freire, Immler, et~al\mbox{.}}{Geretti
  et~al\mbox{.}}{2020}]%
        {geretti2020arch}
\bibfield{author}{\bibinfo{person}{Luca Geretti}, \bibinfo{person}{Julien
  Alexandre~Dit Sandretto}, \bibinfo{person}{Matthias Althoff},
  \bibinfo{person}{Luis Benet}, \bibinfo{person}{Alexandre Chapoutot},
  \bibinfo{person}{Xin Chen}, \bibinfo{person}{Pieter Collins},
  \bibinfo{person}{Marcelo Forets}, \bibinfo{person}{Daniel Freire},
  \bibinfo{person}{Fabian Immler}, {et~al\mbox{.}}}
  \bibinfo{year}{2020}\natexlab{}.
\newblock \showarticletitle{Arch-comp20 category report: Continuous and hybrid
  systems with nonlinear dynamics}.
\newblock \bibinfo{journal}{\emph{EPiC Series in Computing}}
  \bibinfo{volume}{74} (\bibinfo{year}{2020}), \bibinfo{pages}{49--75}.
\newblock


\bibitem[\protect\citeauthoryear{Gr{\"u}new{\"a}lder, Lever, Baldassarre,
  Patterson, Gretton, and Pontil}{Gr{\"u}new{\"a}lder et~al\mbox{.}}{2012}]%
        {grunewalder2012conditional}
\bibfield{author}{\bibinfo{person}{Steffen Gr{\"u}new{\"a}lder},
  \bibinfo{person}{Guy Lever}, \bibinfo{person}{Luca Baldassarre},
  \bibinfo{person}{Sam Patterson}, \bibinfo{person}{Arthur Gretton}, {and}
  \bibinfo{person}{Massimilano Pontil}.} \bibinfo{year}{2012}\natexlab{}.
\newblock \showarticletitle{Conditional mean embeddings as regressors}. In
  \bibinfo{booktitle}{\emph{Proceedings of the 29th International Coference on
  International Conference on Machine Learning}}. \bibinfo{pages}{1803--1810}.
\newblock


\bibitem[\protect\citeauthoryear{Gr\"{u}new\"{a}lder, Lever, Baldassarre,
  Pontil, and Gretton}{Gr\"{u}new\"{a}lder et~al\mbox{.}}{2012}]%
        {grunewalder2012modelling}
\bibfield{author}{\bibinfo{person}{Steffen Gr\"{u}new\"{a}lder},
  \bibinfo{person}{Guy Lever}, \bibinfo{person}{Luca Baldassarre},
  \bibinfo{person}{Massimilano Pontil}, {and} \bibinfo{person}{Arthur
  Gretton}.} \bibinfo{year}{2012}\natexlab{}.
\newblock \showarticletitle{Modelling Transition Dynamics in MDPs with RKHS
  Embeddings}. In \bibinfo{booktitle}{\emph{Proceedings of the 29th
  International Coference on International Conference on Machine Learning}}
  \emph{(\bibinfo{series}{ICML'12})}. \bibinfo{publisher}{Omnipress},
  \bibinfo{address}{Madison, WI, USA}, \bibinfo{pages}{1603–1610}.
\newblock
\showISBNx{9781450312851}


\bibitem[\protect\citeauthoryear{Katz, Huang, Ibeling, Julian, Lazarus, Lim,
  Shah, Thakoor, Wu, Zelji{\'{c}}, Dill, Kochenderfer, and Barrett}{Katz
  et~al\mbox{.}}{2019}]%
        {marabou}
\bibfield{author}{\bibinfo{person}{Guy Katz}, \bibinfo{person}{Derek Huang},
  \bibinfo{person}{Duligur Ibeling}, \bibinfo{person}{Kyle Julian},
  \bibinfo{person}{Christopher Lazarus}, \bibinfo{person}{Rachel Lim},
  \bibinfo{person}{Parth Shah}, \bibinfo{person}{Shantanu Thakoor},
  \bibinfo{person}{Haoze Wu}, \bibinfo{person}{Aleksandar Zelji{\'{c}}},
  \bibinfo{person}{David~L. Dill}, \bibinfo{person}{Mykel Kochenderfer}, {and}
  \bibinfo{person}{Clark Barrett}.} \bibinfo{year}{2019}\natexlab{}.
\newblock \showarticletitle{The Marabou Framework for Verification and Analysis
  of Deep Neural Networks}. In \bibinfo{booktitle}{\emph{Computer Aided
  Verification}}, \bibfield{editor}{\bibinfo{person}{Isil Dillig} {and}
  \bibinfo{person}{Serdar Tasiran}} (Eds.). \bibinfo{publisher}{Springer
  International Publishing}, \bibinfo{address}{Cham},
  \bibinfo{pages}{443--452}.
\newblock


\bibitem[\protect\citeauthoryear{Kingston, Moll, and Kavraki}{Kingston
  et~al\mbox{.}}{2018}]%
        {kingston2018sampling}
\bibfield{author}{\bibinfo{person}{Zachary Kingston}, \bibinfo{person}{Mark
  Moll}, {and} \bibinfo{person}{Lydia~E Kavraki}.}
  \bibinfo{year}{2018}\natexlab{}.
\newblock \showarticletitle{Sampling-based methods for motion planning with
  constraints}.
\newblock \bibinfo{journal}{\emph{Annual review of control, robotics, and
  autonomous systems}}  \bibinfo{volume}{1} (\bibinfo{year}{2018}),
  \bibinfo{pages}{159--185}.
\newblock


\bibitem[\protect\citeauthoryear{Kwiatkowska, Norman, and Parker}{Kwiatkowska
  et~al\mbox{.}}{2011}]%
        {kwiatkowska2011prism}
\bibfield{author}{\bibinfo{person}{Marta Kwiatkowska}, \bibinfo{person}{Gethin
  Norman}, {and} \bibinfo{person}{David Parker}.}
  \bibinfo{year}{2011}\natexlab{}.
\newblock \showarticletitle{PRISM 4.0: Verification of probabilistic real-time
  systems}. In \bibinfo{booktitle}{\emph{International conference on computer
  aided verification}}. Springer, \bibinfo{pages}{585--591}.
\newblock


\bibitem[\protect\citeauthoryear{Lavaei, Khaled, Soudjani, and Zamani}{Lavaei
  et~al\mbox{.}}{2020}]%
        {lavaei2020amytiss}
\bibfield{author}{\bibinfo{person}{Abolfazl Lavaei}, \bibinfo{person}{Mahmoud
  Khaled}, \bibinfo{person}{Sadegh Soudjani}, {and} \bibinfo{person}{Majid
  Zamani}.} \bibinfo{year}{2020}\natexlab{}.
\newblock \showarticletitle{AMYTISS: A Parallelized Tool on Automated
  Controller Synthesis for Large-Scale Stochastic Systems}. In
  \bibinfo{booktitle}{\emph{Proceedings of the 23rd International Conference on
  Hybrid Systems: Computation and Control}} \emph{(\bibinfo{series}{HSCC
  '20})}. \bibinfo{publisher}{Association for Computing Machinery},
  \bibinfo{address}{New York, NY, USA}, Article \bibinfo{articleno}{31},
  \bibinfo{numpages}{2}~pages.
\newblock
\showISBNx{9781450370189}


\bibitem[\protect\citeauthoryear{Lesser, Oishi, and Erwin}{Lesser
  et~al\mbox{.}}{2013}]%
        {lesser2013stochastic}
\bibfield{author}{\bibinfo{person}{Kendra Lesser}, \bibinfo{person}{Meeko
  Oishi}, {and} \bibinfo{person}{R~Scott Erwin}.}
  \bibinfo{year}{2013}\natexlab{}.
\newblock \showarticletitle{Stochastic reachability for control of spacecraft
  relative motion}. In \bibinfo{booktitle}{\emph{52nd IEEE Conference on
  Decision and Control}}. IEEE, \bibinfo{pages}{4705--4712}.
\newblock


\bibitem[\protect\citeauthoryear{Lever and Stafford}{Lever and
  Stafford}{2015}]%
        {lever2015modelling}
\bibfield{author}{\bibinfo{person}{Guy Lever} {and} \bibinfo{person}{Ronnie
  Stafford}.} \bibinfo{year}{2015}\natexlab{}.
\newblock \showarticletitle{{Modelling Policies in MDPs in Reproducing Kernel
  Hilbert Space}}. In \bibinfo{booktitle}{\emph{Proceedings of the Eighteenth
  International Conference on Artificial Intelligence and Statistics}}
  \emph{(\bibinfo{series}{Proceedings of Machine Learning Research})},
  \bibfield{editor}{\bibinfo{person}{Guy Lebanon} {and}
  \bibinfo{person}{S.~V.~N. Vishwanathan}} (Eds.), Vol.~\bibinfo{volume}{38}.
  \bibinfo{publisher}{PMLR}, \bibinfo{address}{San Diego, California, USA},
  \bibinfo{pages}{590--598}.
\newblock


\bibitem[\protect\citeauthoryear{Marinho, Boots, Dragan, Byravan, Gordon, and
  Srinivasa}{Marinho et~al\mbox{.}}{2016}]%
        {marinho2016functional}
\bibfield{author}{\bibinfo{person}{Zita Marinho}, \bibinfo{person}{Byron
  Boots}, \bibinfo{person}{Anca Dragan}, \bibinfo{person}{Arunkumar Byravan},
  \bibinfo{person}{Geoffrey~J. Gordon}, {and} \bibinfo{person}{Siddhartha
  Srinivasa}.} \bibinfo{year}{2016}\natexlab{}.
\newblock \showarticletitle{Functional Gradient Motion Planning in Reproducing
  Kernel Hilbert Spaces}. In \bibinfo{booktitle}{\emph{Proceedings of Robotics:
  Science and Systems}}. \bibinfo{address}{AnnArbor, Michigan}.
\newblock
\urldef\tempurl%
\url{https://doi.org/10.15607/RSS.2016.XII.046}
\showDOI{\tempurl}


\bibitem[\protect\citeauthoryear{Micchelli and Pontil}{Micchelli and
  Pontil}{2005}]%
        {micchelli2005learning}
\bibfield{author}{\bibinfo{person}{Charles~A Micchelli} {and}
  \bibinfo{person}{Massimiliano Pontil}.} \bibinfo{year}{2005}\natexlab{}.
\newblock \showarticletitle{On learning vector-valued functions}.
\newblock \bibinfo{journal}{\emph{Neural computation}} \bibinfo{volume}{17},
  \bibinfo{number}{1} (\bibinfo{year}{2005}), \bibinfo{pages}{177--204}.
\newblock


\bibitem[\protect\citeauthoryear{Park and Muandet}{Park and Muandet}{2020}]%
        {park2020measure}
\bibfield{author}{\bibinfo{person}{Junhyung Park} {and}
  \bibinfo{person}{Krikamol Muandet}.} \bibinfo{year}{2020}\natexlab{}.
\newblock \showarticletitle{A measure-theoretic approach to kernel conditional
  mean embeddings}.
\newblock \bibinfo{journal}{\emph{Advances in Neural Information Processing
  Systems}}  \bibinfo{volume}{33} (\bibinfo{year}{2020}).
\newblock


\bibitem[\protect\citeauthoryear{Paszke, Gross, Massa, Lerer, Bradbury, Chanan,
  Killeen, Lin, Gimelshein, Antiga, Desmaison, Kopf, Yang, DeVito, Raison,
  Tejani, Chilamkurthy, Steiner, Fang, Bai, and Chintala}{Paszke
  et~al\mbox{.}}{2019}]%
        {NEURIPS2019_9015}
\bibfield{author}{\bibinfo{person}{Adam Paszke}, \bibinfo{person}{Sam Gross},
  \bibinfo{person}{Francisco Massa}, \bibinfo{person}{Adam Lerer},
  \bibinfo{person}{James Bradbury}, \bibinfo{person}{Gregory Chanan},
  \bibinfo{person}{Trevor Killeen}, \bibinfo{person}{Zeming Lin},
  \bibinfo{person}{Natalia Gimelshein}, \bibinfo{person}{Luca Antiga},
  \bibinfo{person}{Alban Desmaison}, \bibinfo{person}{Andreas Kopf},
  \bibinfo{person}{Edward Yang}, \bibinfo{person}{Zachary DeVito},
  \bibinfo{person}{Martin Raison}, \bibinfo{person}{Alykhan Tejani},
  \bibinfo{person}{Sasank Chilamkurthy}, \bibinfo{person}{Benoit Steiner},
  \bibinfo{person}{Lu Fang}, \bibinfo{person}{Junjie Bai}, {and}
  \bibinfo{person}{Soumith Chintala}.} \bibinfo{year}{2019}\natexlab{}.
\newblock \showarticletitle{PyTorch: An Imperative Style, High-Performance Deep
  Learning Library}.
\newblock In \bibinfo{booktitle}{\emph{Advances in Neural Information
  Processing Systems 32}}, \bibfield{editor}{\bibinfo{person}{H.~Wallach},
  \bibinfo{person}{H.~Larochelle}, \bibinfo{person}{A.~Beygelzimer},
  \bibinfo{person}{F.~d\textquotesingle Alch\'{e}-Buc},
  \bibinfo{person}{E.~Fox}, {and} \bibinfo{person}{R.~Garnett}} (Eds.).
  \bibinfo{publisher}{Curran Associates, Inc.}, \bibinfo{pages}{8024--8035}.
\newblock


\bibitem[\protect\citeauthoryear{Pedregosa, Varoquaux, Gramfort, Michel,
  Thirion, Grisel, Blondel, Prettenhofer, Weiss, Dubourg, Vanderplas, Passos,
  Cournapeau, Brucher, Perrot, and Duchesnay}{Pedregosa et~al\mbox{.}}{2011}]%
        {scikit-learn}
\bibfield{author}{\bibinfo{person}{F. Pedregosa}, \bibinfo{person}{G.
  Varoquaux}, \bibinfo{person}{A. Gramfort}, \bibinfo{person}{V. Michel},
  \bibinfo{person}{B. Thirion}, \bibinfo{person}{O. Grisel},
  \bibinfo{person}{M. Blondel}, \bibinfo{person}{P. Prettenhofer},
  \bibinfo{person}{R. Weiss}, \bibinfo{person}{V. Dubourg}, \bibinfo{person}{J.
  Vanderplas}, \bibinfo{person}{A. Passos}, \bibinfo{person}{D. Cournapeau},
  \bibinfo{person}{M. Brucher}, \bibinfo{person}{M. Perrot}, {and}
  \bibinfo{person}{E. Duchesnay}.} \bibinfo{year}{2011}\natexlab{}.
\newblock \showarticletitle{Scikit-learn: Machine Learning in {P}ython}.
\newblock \bibinfo{journal}{\emph{Journal of Machine Learning Research}}
  \bibinfo{volume}{12} (\bibinfo{year}{2011}), \bibinfo{pages}{2825--2830}.
\newblock


\bibitem[\protect\citeauthoryear{Rahimi and Recht}{Rahimi and Recht}{2007}]%
        {rahimi2007random}
\bibfield{author}{\bibinfo{person}{Ali Rahimi} {and} \bibinfo{person}{Benjamin
  Recht}.} \bibinfo{year}{2007}\natexlab{}.
\newblock \showarticletitle{Random Features for Large-Scale Kernel Machines}.
  In \bibinfo{booktitle}{\emph{Advances in Neural Information Processing
  Systems}}, \bibfield{editor}{\bibinfo{person}{J.~Platt},
  \bibinfo{person}{D.~Koller}, \bibinfo{person}{Y.~Singer}, {and}
  \bibinfo{person}{S.~Roweis}} (Eds.), Vol.~\bibinfo{volume}{20}.
  \bibinfo{publisher}{Curran Associates, Inc.}
\newblock
\urldef\tempurl%
\url{https://proceedings.neurips.cc/paper/2007/file/013a006f03dbc5392effeb8f18fda755-Paper.pdf}
\showURL{%
\tempurl}


\bibitem[\protect\citeauthoryear{Rasmussen and Williams}{Rasmussen and
  Williams}{2006}]%
        {rasmussen2006gaussian}
\bibfield{author}{\bibinfo{person}{Carl~Edward Rasmussen} {and}
  \bibinfo{person}{Chris Williams}.} \bibinfo{year}{2006}\natexlab{}.
\newblock \bibinfo{booktitle}{\emph{Gaussian Processes for Machine Learning}}.
\newblock \bibinfo{publisher}{MIT Press}.
\newblock


\bibitem[\protect\citeauthoryear{Ray, Achiam, and Amodei}{Ray
  et~al\mbox{.}}{2019}]%
        {ray2019benchmarking}
\bibfield{author}{\bibinfo{person}{Alex Ray}, \bibinfo{person}{Joshua Achiam},
  {and} \bibinfo{person}{Dario Amodei}.} \bibinfo{year}{2019}\natexlab{}.
\newblock \showarticletitle{Benchmarking safe exploration in deep reinforcement
  learning}.
\newblock  (\bibinfo{year}{2019}).
\newblock


\bibitem[\protect\citeauthoryear{Reddy, Dragan, Levine, Legg, and Leike}{Reddy
  et~al\mbox{.}}{2020}]%
        {reddy2020learning}
\bibfield{author}{\bibinfo{person}{Siddharth Reddy}, \bibinfo{person}{Anca
  Dragan}, \bibinfo{person}{Sergey Levine}, \bibinfo{person}{Shane Legg}, {and}
  \bibinfo{person}{Jan Leike}.} \bibinfo{year}{2020}\natexlab{}.
\newblock \showarticletitle{Learning human objectives by evaluating
  hypothetical behavior}. In \bibinfo{booktitle}{\emph{International Conference
  on Machine Learning}}. PMLR, \bibinfo{pages}{8020--8029}.
\newblock


\bibitem[\protect\citeauthoryear{Rosolia and Borrelli}{Rosolia and
  Borrelli}{2017}]%
        {rosolia2017learning}
\bibfield{author}{\bibinfo{person}{Ugo Rosolia} {and}
  \bibinfo{person}{Francesco Borrelli}.} \bibinfo{year}{2017}\natexlab{}.
\newblock \showarticletitle{Learning model predictive control for iterative
  tasks. a data-driven control framework}.
\newblock \bibinfo{journal}{\emph{IEEE Trans. Automat. Control}}
  \bibinfo{volume}{63}, \bibinfo{number}{7} (\bibinfo{year}{2017}),
  \bibinfo{pages}{1883--1896}.
\newblock


\bibitem[\protect\citeauthoryear{Rosolia and Borrelli}{Rosolia and
  Borrelli}{2019}]%
        {rosolia2019sample}
\bibfield{author}{\bibinfo{person}{Ugo Rosolia} {and}
  \bibinfo{person}{Francesco Borrelli}.} \bibinfo{year}{2019}\natexlab{}.
\newblock \showarticletitle{Sample-based learning model predictive control for
  linear uncertain systems}. In \bibinfo{booktitle}{\emph{2019 IEEE 58th
  Conference on Decision and Control (CDC)}}. IEEE,
  \bibinfo{pages}{2702--2707}.
\newblock


\bibitem[\protect\citeauthoryear{Sch{\"o}lkopf, Smola, Bach,
  et~al\mbox{.}}{Sch{\"o}lkopf et~al\mbox{.}}{2002}]%
        {scholkopf2002learning}
\bibfield{author}{\bibinfo{person}{Bernhard Sch{\"o}lkopf},
  \bibinfo{person}{Alexander~J Smola}, \bibinfo{person}{Francis Bach},
  {et~al\mbox{.}}} \bibinfo{year}{2002}\natexlab{}.
\newblock \bibinfo{booktitle}{\emph{Learning with kernels: support vector
  machines, regularization, optimization, and beyond}}.
\newblock \bibinfo{publisher}{MIT press}.
\newblock


\bibitem[\protect\citeauthoryear{Shmarov and Zuliani}{Shmarov and
  Zuliani}{2015}]%
        {shmarov2015probreach}
\bibfield{author}{\bibinfo{person}{Fedor Shmarov} {and} \bibinfo{person}{Paolo
  Zuliani}.} \bibinfo{year}{2015}\natexlab{}.
\newblock \showarticletitle{ProbReach: Verified Probabilistic
  Delta-Reachability for Stochastic Hybrid Systems}. In
  \bibinfo{booktitle}{\emph{Proceedings of the 18th International Conference on
  Hybrid Systems: Computation and Control}} \emph{(\bibinfo{series}{HSCC
  '15})}. \bibinfo{publisher}{Association for Computing Machinery},
  \bibinfo{address}{New York, NY, USA}, \bibinfo{pages}{134–139}.
\newblock
\showISBNx{9781450334334}
\urldef\tempurl%
\url{https://doi.org/10.1145/2728606.2728625}
\showDOI{\tempurl}


\bibitem[\protect\citeauthoryear{Smola, Gretton, Song, and Sch{\"o}lkopf}{Smola
  et~al\mbox{.}}{2007}]%
        {smola2007hilbert}
\bibfield{author}{\bibinfo{person}{Alex Smola}, \bibinfo{person}{Arthur
  Gretton}, \bibinfo{person}{Le Song}, {and} \bibinfo{person}{Bernhard
  Sch{\"o}lkopf}.} \bibinfo{year}{2007}\natexlab{}.
\newblock \showarticletitle{A Hilbert space embedding for distributions}. In
  \bibinfo{booktitle}{\emph{International Conference on Algorithmic Learning
  Theory}}. Springer, \bibinfo{pages}{13--31}.
\newblock


\bibitem[\protect\citeauthoryear{Song, Boots, Siddiqi, Gordon, and Smola}{Song
  et~al\mbox{.}}{2010a}]%
        {song2010hilbert}
\bibfield{author}{\bibinfo{person}{Le Song}, \bibinfo{person}{Byron Boots},
  \bibinfo{person}{Sajid~M. Siddiqi}, \bibinfo{person}{Geoffrey Gordon}, {and}
  \bibinfo{person}{Alex Smola}.} \bibinfo{year}{2010}\natexlab{a}.
\newblock \showarticletitle{Hilbert Space Embeddings of Hidden Markov Models}.
  In \bibinfo{booktitle}{\emph{Proceedings of the 27th International Conference
  on International Conference on Machine Learning}}
  \emph{(\bibinfo{series}{ICML'10})}. \bibinfo{publisher}{Omnipress},
  \bibinfo{address}{Madison, WI, USA}, \bibinfo{pages}{991–998}.
\newblock
\showISBNx{9781605589077}


\bibitem[\protect\citeauthoryear{Song, Gretton, and Guestrin}{Song
  et~al\mbox{.}}{2010b}]%
        {song2010nonparametric}
\bibfield{author}{\bibinfo{person}{Le Song}, \bibinfo{person}{Arthur Gretton},
  {and} \bibinfo{person}{Carlos Guestrin}.} \bibinfo{year}{2010}\natexlab{b}.
\newblock \showarticletitle{Nonparametric Tree Graphical Models}. In
  \bibinfo{booktitle}{\emph{Proceedings of the Thirteenth International
  Conference on Artificial Intelligence and Statistics}}
  \emph{(\bibinfo{series}{Proceedings of Machine Learning Research})},
  \bibfield{editor}{\bibinfo{person}{Yee~Whye Teh} {and} \bibinfo{person}{Mike
  Titterington}} (Eds.), Vol.~\bibinfo{volume}{9}. \bibinfo{publisher}{PMLR},
  \bibinfo{address}{Chia Laguna Resort, Sardinia, Italy},
  \bibinfo{pages}{765--772}.
\newblock


\bibitem[\protect\citeauthoryear{Song, Huang, Smola, and Fukumizu}{Song
  et~al\mbox{.}}{2009}]%
        {song2009hilbert}
\bibfield{author}{\bibinfo{person}{Le Song}, \bibinfo{person}{Jonathan Huang},
  \bibinfo{person}{Alex Smola}, {and} \bibinfo{person}{Kenji Fukumizu}.}
  \bibinfo{year}{2009}\natexlab{}.
\newblock \showarticletitle{Hilbert space embeddings of conditional
  distributions with applications to dynamical systems}. In
  \bibinfo{booktitle}{\emph{Proceedings of the 26th Annual International
  Conference on Machine Learning}}. \bibinfo{pages}{961--968}.
\newblock


\bibitem[\protect\citeauthoryear{Soudjani, Gevaerts, and Abate}{Soudjani
  et~al\mbox{.}}{2015}]%
        {soudjani2014faust}
\bibfield{author}{\bibinfo{person}{Sadegh Esmaeil~Zadeh Soudjani},
  \bibinfo{person}{Caspar Gevaerts}, {and} \bibinfo{person}{Alessandro Abate}.}
  \bibinfo{year}{2015}\natexlab{}.
\newblock \showarticletitle{{FAUST 2 : Formal Abstractions of Uncountable-STate
  STochastic Processes}}. In \bibinfo{booktitle}{\emph{{International
  Conference on Tools and Algorithms for the Construction and Analysis of
  Systems}}}, Vol.~\bibinfo{volume}{9035}. \bibinfo{publisher}{{Springer
  International Publishing}}, \bibinfo{pages}{272--286}.
\newblock


\bibitem[\protect\citeauthoryear{Steinwart and Christmann}{Steinwart and
  Christmann}{2008}]%
        {steinwart2008support}
\bibfield{author}{\bibinfo{person}{Ingo Steinwart} {and}
  \bibinfo{person}{Andreas Christmann}.} \bibinfo{year}{2008}\natexlab{}.
\newblock \bibinfo{booktitle}{\emph{Support Vector Machines}}.
\newblock \bibinfo{publisher}{Springer Publishing Company, Incorporated}.
\newblock


\bibitem[\protect\citeauthoryear{Summers and Lygeros}{Summers and
  Lygeros}{2010}]%
        {summers2010verification}
\bibfield{author}{\bibinfo{person}{Sean Summers} {and} \bibinfo{person}{John
  Lygeros}.} \bibinfo{year}{2010}\natexlab{}.
\newblock \showarticletitle{Verification of discrete time stochastic hybrid
  systems: A stochastic reach-avoid decision problem}.
\newblock \bibinfo{journal}{\emph{Automatica}} \bibinfo{volume}{46},
  \bibinfo{number}{12} (\bibinfo{year}{2010}), \bibinfo{pages}{1951--1961}.
\newblock


\bibitem[\protect\citeauthoryear{Thorpe and Oishi}{Thorpe and Oishi}{2020}]%
        {thorpe2020model}
\bibfield{author}{\bibinfo{person}{Adam~J. Thorpe} {and} \bibinfo{person}{Meeko
  M.~K. Oishi}.} \bibinfo{year}{2020}\natexlab{}.
\newblock \showarticletitle{Model-Free Stochastic Reachability Using Kernel
  Distribution Embeddings}.
\newblock \bibinfo{journal}{\emph{IEEE Control Systems Letters}}
  \bibinfo{volume}{4}, \bibinfo{number}{2} (\bibinfo{year}{2020}),
  \bibinfo{pages}{512--517}.
\newblock


\bibitem[\protect\citeauthoryear{Thorpe and Oishi}{Thorpe and Oishi}{2021}]%
        {thorpe2021stochastic}
\bibfield{author}{\bibinfo{person}{Adam~J. Thorpe} {and} \bibinfo{person}{Meeko
  M.~K. Oishi}.} \bibinfo{year}{2021}\natexlab{}.
\newblock \showarticletitle{Stochastic Optimal Control via Hilbert Space
  Embeddings of Distributions}. In \bibinfo{booktitle}{\emph{2021 60th IEEE
  Conference on Decision and Control (CDC)}}. \bibinfo{pages}{904--911}.
\newblock
\urldef\tempurl%
\url{https://doi.org/10.1109/CDC45484.2021.9682801}
\showDOI{\tempurl}


\bibitem[\protect\citeauthoryear{Thorpe, Ortiz, and Oishi}{Thorpe
  et~al\mbox{.}}{2021a}]%
        {thorpe2021learning}
\bibfield{author}{\bibinfo{person}{Adam~J. Thorpe}, \bibinfo{person}{Kendric~R.
  Ortiz}, {and} \bibinfo{person}{Meeko M.~K. Oishi}.}
  \bibinfo{year}{2021}\natexlab{a}.
\newblock \showarticletitle{Learning Approximate Forward Reachable Sets Using
  Separating Kernels}. In \bibinfo{booktitle}{\emph{Proceedings of the 3rd
  Conference on Learning for Dynamics and Control}}
  \emph{(\bibinfo{series}{Proceedings of Machine Learning Research})},
  \bibfield{editor}{\bibinfo{person}{Ali Jadbabaie}, \bibinfo{person}{John
  Lygeros}, \bibinfo{person}{George~J. Pappas}, \bibinfo{person}{Pablo
  A.~Parrilo}, \bibinfo{person}{Benjamin Recht}, \bibinfo{person}{Claire~J.
  Tomlin}, {and} \bibinfo{person}{Melanie~N. Zeilinger}} (Eds.),
  Vol.~\bibinfo{volume}{144}. \bibinfo{publisher}{PMLR},
  \bibinfo{pages}{201--212}.
\newblock


\bibitem[\protect\citeauthoryear{Thorpe, Ortiz, and Oishi}{Thorpe
  et~al\mbox{.}}{2021b}]%
        {thorpe2020sreachtools}
\bibfield{author}{\bibinfo{person}{Adam~J. Thorpe}, \bibinfo{person}{Kendric~R.
  Ortiz}, {and} \bibinfo{person}{Meeko M.~K. Oishi}.}
  \bibinfo{year}{2021}\natexlab{b}.
\newblock \showarticletitle{SReachTools Kernel Module: Data-Driven Stochastic
  Reachability Using Hilbert Space Embeddings of Distributions}. In
  \bibinfo{booktitle}{\emph{2021 60th IEEE Conference on Decision and Control
  (CDC)}}. \bibinfo{pages}{5073--5079}.
\newblock
\urldef\tempurl%
\url{https://doi.org/10.1109/CDC45484.2021.9683169}
\showDOI{\tempurl}


\bibitem[\protect\citeauthoryear{Thorpe, Sivaramakrishnan, and Oishi}{Thorpe
  et~al\mbox{.}}{2021c}]%
        {thorpe2021approximate}
\bibfield{author}{\bibinfo{person}{Adam~J. Thorpe}, \bibinfo{person}{Vignesh
  Sivaramakrishnan}, {and} \bibinfo{person}{Meeko M.~K. Oishi}.}
  \bibinfo{year}{2021}\natexlab{c}.
\newblock \showarticletitle{Approximate Stochastic Reachability for High
  Dimensional Systems}. In \bibinfo{booktitle}{\emph{2021 American Control
  Conference (ACC)}}. \bibinfo{pages}{1287--1293}.
\newblock


\bibitem[\protect\citeauthoryear{Tran, Musau, Lopez, Yang, Nguyen, Xiang, and
  Johnson}{Tran et~al\mbox{.}}{2020}]%
        {nnv}
\bibfield{author}{\bibinfo{person}{Hoang-Dung Tran}, \bibinfo{person}{Patrick
  Musau}, \bibinfo{person}{Diego~Manzanas Lopez}, \bibinfo{person}{Xiaodong
  Yang}, \bibinfo{person}{Luan~Viet Nguyen}, \bibinfo{person}{Weiming Xiang},
  {and} \bibinfo{person}{Taylor Johnson}.} \bibinfo{year}{2020}\natexlab{}.
\newblock \showarticletitle{{NNV}: A Tool for Verification of Deep Neural
  Networks and Learning-Enabled Autonomous Cyber-Physical Systems}. In
  \bibinfo{booktitle}{\emph{International Conference on Computer-Aided
  Verification}}.
\newblock


\bibitem[\protect\citeauthoryear{Vinod, Gleason, and Oishi}{Vinod
  et~al\mbox{.}}{2019}]%
        {sreachtools}
\bibfield{author}{\bibinfo{person}{Abraham Vinod}, \bibinfo{person}{Joseph
  Gleason}, {and} \bibinfo{person}{Meeko Oishi}.}
  \bibinfo{year}{2019}\natexlab{}.
\newblock \showarticletitle{SReachTools: a MATLAB stochastic reachability
  toolbox}. In \bibinfo{booktitle}{\emph{International Conference on Hybrid
  Systems: Computation and Control}}. \bibinfo{publisher}{ACM},
  \bibinfo{pages}{33–38}.
\newblock
\showISBNx{978-1-4503-6282-5}


\bibitem[\protect\citeauthoryear{Zhu, Jitkrittum, Diehl, and Sch{\"o}lkopf}{Zhu
  et~al\mbox{.}}{2021}]%
        {zhu2021kernel}
\bibfield{author}{\bibinfo{person}{Jia-Jie Zhu}, \bibinfo{person}{Wittawat
  Jitkrittum}, \bibinfo{person}{Moritz Diehl}, {and} \bibinfo{person}{Bernhard
  Sch{\"o}lkopf}.} \bibinfo{year}{2021}\natexlab{}.
\newblock \showarticletitle{Kernel Distributionally Robust Optimization:
  Generalized Duality Theorem and Stochastic Approximation}. In
  \bibinfo{booktitle}{\emph{Proceedings of The 24th International Conference on
  Artificial Intelligence and Statistics}} \emph{(\bibinfo{series}{Proceedings
  of Machine Learning Research})}, \bibfield{editor}{\bibinfo{person}{Arindam
  Banerjee} {and} \bibinfo{person}{Kenji Fukumizu}} (Eds.),
  Vol.~\bibinfo{volume}{130}. \bibinfo{publisher}{PMLR},
  \bibinfo{pages}{280--288}.
\newblock


\end{thebibliography}

\end{document}